\newif\iffinal
\newif\ifarxiv
\arxivtrue
\finaltrue

\documentclass[letterpaper,twocolumn,10pt]{article}
\usepackage[breaklinks,backref=page]{hyperref}
\usepackage{usenix}

\usepackage{hyperref}

\usepackage{microtype}
\usepackage{tikz}
\usepackage{subfig}
\usepackage{multirow}
\usepackage{booktabs}
\usepackage{graphicx}
\usepackage{xspace}
\usepackage{algorithm}
\usepackage[noend]{algpseudocode}
\usepackage{balance}
\usepackage{amsmath}
\usepackage{mathtools}
\usepackage{amsfonts}
\usepackage{amssymb}
\usepackage{amsthm}
\usepackage{makecell}
\usepackage{threeparttable}
\usepackage[available, functional, reproduced]{usenixbadges}

\usepackage[nosort]{cleveref} 

\newmuskip\normalthickmuskip
\newmuskip\normalthinmuskip
\AtBeginDocument{%
  \normalthickmuskip=\thickmuskip
  \normalthinmuskip=\thinmuskip
}
\everymath{%
  \thickmuskip=\muexpr\normalthickmuskip-(1\normalthinmuskip plus 1\normalthinmuskip)\relax
}
\everydisplay{\thickmuskip=\normalthickmuskip}
\let\texdisplaystyle\displaystyle
\renewcommand{\displaystyle}{\texdisplaystyle\the\everydisplay}

\algrenewcommand\algorithmicindent{0.5em}
\algrenewcommand\algorithmicrequire{\textbf{Input:}}
\algrenewcommand\algorithmicensure{\textbf{Output:}}

\newcommand{\ZJ}[1] {\textcolor{blue}{[ZJ: #1]}}
\newcommand{\MW}[1]{\textcolor{green}{[MW: #1]}}
\newcommand{\GO}[1]{\textcolor{olive}{GO: #1}}
\newcommand{\oded}[1]{\textcolor{cyan}{OP: #1}}

\theoremstyle{definition}
\newtheorem{definition}{Definition}[section]

\renewcommand{\ZJ}[1]{\ifdim\lastskip>0pt\ignorespaces\fi}
\renewcommand{\MW}[1]{\ifdim\lastskip>0pt\ignorespaces\fi}
\renewcommand{\GO}[1]{\ifdim\lastskip>0pt\ignorespaces\fi}
\renewcommand{\oded}[1]{\ifdim\lastskip>0pt\ignorespaces\fi}

\newtheorem{theorem}{Theorem}

\ifarxiv
\newcommand{\Sys}{Mirage\xspace}
\newcommand{\sys}{Mirage\xspace}
\else
\newcommand{\Sys}{MISO\xspace}
\newcommand{\sys}{MISO\xspace}
\fi
\newcommand{\lax}{\textproc{Lax}\xspace}
\newcommand{\llama}{LLaMA\xspace}
\newcommand{\commentout}[1]{}
\algnewcommand{\LeftComment}[1]{\Statex \(\triangleright\) #1}
\newcommand{\er}[1]{\mbox{\rm\em #1}}

\newcommand{\removed}[1]{}

\iffinal
\newcommand{\captionvspace}{-0.5em}
\else
\newcommand{\captionvspace}{-1em}
\fi

\crefname{part}{\S}{\S\S}
\crefname{chapter}{\S}{\S\S}
\crefname{section}{\S}{\S\S}
\crefname{subsection}{\S}{\S\S}

\newcommand{\imap}{\er{imap}\xspace}
\newcommand{\omap}{\er{omap}\xspace}
\newcommand{\fmap}{\er{fmap}\xspace}
\newcommand{\graph}{$\mu$Graph\xspace}
\newcommand{\graphs}{$\mu$Graphs\xspace}

\newcommand{\GF}[1]{G_{\mathsf{#1}}}
\newcommand{\ngraph}{\GF{}}
\newcommand{\kngraph}{\GF{K}}
\newcommand{\tbgraph}{\GF{B}}
\newcommand{\refgraph}{\GF{ref}}

\newcommand{\vi}[1]{\mathit{#1}}
\newcommand{\func}[1]{\mathrm{#1}}
\newcommand{\eadd}{\mathsf{add}}
\newcommand{\emul}{\mathsf{mul}}
\newcommand{\ediv}{\mathsf{div}}
\newcommand{\eexp}{\mathsf{exp}}
\newcommand{\ered}{\mathsf{sum}}
\newcommand{\esqrt}{\mathsf{sqrt}}
\newcommand{\esilu}{\mathsf{silu}}
\newcommand{\subexpr}{\mathsf{subexpr}}
\newcommand{\expr}{\mathrm{E}}
\newcommand{\prefix}{prefix\xspace}
\newcommand{\Aeq}{A_\text{eq}}
\newcommand{\Asub}{A_\text{sub}}
\begin{document}

\ifarxiv
\title{\sys: A Multi-Level Superoptimizer for Tensor Programs}
\author{\rm{Mengdi Wu} \hspace{1.2em} \rm{Xinhao Cheng} \hspace{1.2em} Shengyu Liu$^{\dag}$ \hspace{1.2em} Chunan Shi$^\dag$ \hspace{1.2em} Jianan Ji\\
 \rm{Man Kit Ao} \hspace{1.2em} Praveen Velliengiri$^{\ddag}$ \hspace{1.2em} Xupeng Miao$^{\sharp}$ \hspace{1.2em} Oded Padon$^{\diamond}$ \hspace{1.2em} Zhihao Jia \\
\\
Carnegie Mellon University \hspace{1.2em} Peking University$^\dag$ \\
Pennsylvania State University$^\ddag$ \hspace{1.2em} Purdue University$^{\sharp}$ \hspace{1.2em} Weizmann Institute of Science$^{\diamond}$
}
\maketitle

\else

\twocolumn[\begin{@twocolumnfalse}

\begin{centering}
{\Large \bf \sys: A Multi-Level Superoptimizer for Tensor Programs\\}
\vspace{0.15cm}
OSDI'25 Submission \#674, 12 pages

\end{centering}

\vspace{\baselineskip}

\end{@twocolumnfalse}]

\renewcommand{\paragraph}[1]{{\vspace{.2em}\noindent\textbf{#1}~~}}

\fi
\pagestyle{empty}

\begin{abstract}
We introduce \sys, the first multi-level superoptimizer for tensor programs. 
A key idea in \sys is \graphs, a uniform representation of tensor programs at the kernel, thread block, and thread levels of the GPU compute hierarchy.
\graphs enable \sys to discover novel optimizations that combine algebraic transformations, schedule transformations, and generation of new custom kernels.
To navigate the large search space, \sys introduces a pruning technique based on abstraction that significantly reduces the search space and provides a certain optimality guarantee.
To ensure that the optimized \graph is equivalent to the input program, \sys introduces a probabilistic equivalence verification procedure with strong theoretical guarantees.
Our evaluation shows that \sys significantly outperforms existing approaches even for DNNs that are widely used and heavily optimized.
\iffinal
\sys is publicly available at \url{https://github.com/mirage-project/mirage}.
\fi 
\end{abstract}





\section{Introduction}
\label{sec:intro}

Enabling high-performance execution of deep neural networks (DNNs) on GPUs is critical for modern ML applications. 
Today's DNN frameworks generally specify DNN computation using tensor programs, which are directed acyclic graphs whose nodes and edges represent tensor algebra operators (e.g., matrix multiplication) and tensors (i.e., $n$-dimensional arrays) shared between operators. 

To optimize an input tensor program, existing frameworks (e.g., PyTorch~\cite{pytorch} and TensorFlow~\cite{Tensorflow}) use manually designed rules to map the tensor program to expert-written GPU kernels.
These approaches generally require extensive engineering efforts to design and implement optimization rules, and they may miss certain optimization opportunities.
To address these challenges, recent work has introduced {\em automated} approaches that optimize tensor programs by searching over a comprehensive space of program transformations and applying them based on their performance on target GPUs.
These approaches generally fall into two categories.

The first category of work, including Halide~\cite{halide}, TVM~\cite{tvm}, and Ansor~\cite{ansor}, is motivated by the idea of algorithm and schedule separation\footnote{In the schedule optimization literature, an algorithm describes what to compute in a kernel and a schedule specifies how to compute the kernel.} introduced in Halide and optimizes the {\em schedule} of a tensor program while fixing the algorithm.
For a given algorithm, these optimizers automatically generate performant kernels by searching for possible strategies to execute the kernel on the target hardware.
However, due to the linear algebra nature of DNNs, a tensor program can be represented by a wide spectrum of mathematically equivalent algorithms. Existing schedule-based optimizers only consider kernels whose algorithms are manually specified by users, resulting in missed optimization opportunities.

The second category of work, including TASO, Grappler, Tensat, and PET, considers {\em algebraic transformations}, which exploit mathematical equivalence among different algorithms for a tensor program~\cite{TASO, tensorflow_grappler, yang2021equality, wang2021pet}. 
Examples of algebraic transformations include (1) converting one linear algebra operator into another, such as transforming a convolution to a matrix multiplication; (2) fusing multiple operators to reduce memory access and kernel overhead; and (3) reorganizing operators based on commutativity, associativity, and distributivity.
%
These optimizers perform algebraic transformations at the algorithm level and require programmers to manually specify the set of available operators and their implementations.
They are thus limited by the performance of the provided kernels.



All existing automated optimization approaches, from both categories, still require programmers to manually specify a set of kernels (each defined by a tensor function), and then explore the search space of algebraic {\em or} schedule transformations.
%
However, some advanced performance optimizations require coordinated transformations across the kernel, thread block, and thread levels of the GPU compute hierarchy, and involve introducing completely new kernel computations (e.g., a custom kernel that decomposes standard kernels and fuses only certain computations).
Such optimizations are not included in the search space of existing automated methods and must still be implemented manually.

One such example is FlashAttention~\cite{dao2023flash} (see \S\ref{subsec:benchmark_results} for details), which optimizes attention~\cite{transformers} on GPUs by reordering operators at the algorithm level (algebraic transformations), reorganizing the computation across GPU kernels (yielding new custom kernels), and adapting the parallelization strategy of each kernel to the GPU architecture (schedule transformations).
The transformations required for this example cannot be automatically discovered by existing frameworks and must therefore be implemented manually.
%
%
%
An implementation of FlashAttention in Triton~\cite{tillet2019triton}, a widely used tensor program optimizer, contains more than 700 lines of code~\cite{triton-flashattention}.



We present \sys, the first {\em multi-level superoptimizer} for tensor programs.
\sys automatically discovers and verifies sophisticated optimizations of tensor programs that require joint optimization of algebraic transformations, schedule transformations, and the discovery of new custom kernels. 

A key idea in \sys is \graphs, a {\em hierarchical graph representation} that specifies tensor programs across multiple levels of the GPU compute hierarchy.
By uniformly treating the kernel, thread block, and thread levels, \graphs can capture both algebraic and schedule transformations across these levels.
Moreover, optimizing a \graph can introduce new custom kernels, which go beyond both algebraic and schedule transformations.
%
%
For example, \sys automatically discovers the \graphs representing FlashAttention~\cite{dao2023flash} and its inference variant FlashDecoding~\cite{tri2023flashdecoding} as well as other \graphs that outperform these manually designed kernels by up to 2.2$\times$ for certain use cases.
Most of these optimizations discovered by \sys are outside the search space of existing methods.

\begin{figure}
    \centering
    \includegraphics[scale=0.48]{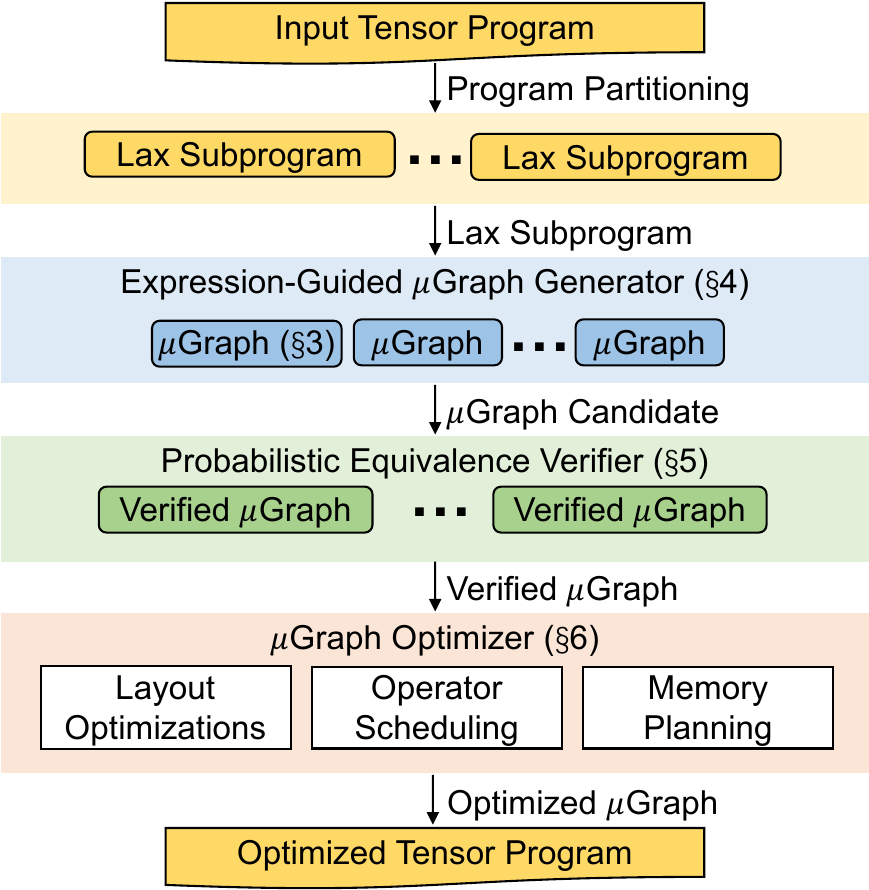}
    \vspace{\captionvspace}
    \caption{An overview of \sys.}
    \vspace{-1em}
    \label{fig:overview}
\end{figure}

\Cref{fig:overview} shows an overview of \sys.
\sys first splits an input tensor program into subprograms that fall into the restricted \lax fragment.
The \lax fragment, formally defined in \S\ref{sec:verify}, includes multi-linear operators such as matrix multiplication and convolution, division (useful for normalizations), and limited exponentiation (useful for activations).
%
Partitioning a tensor program into \lax subprograms reduces the optimization search space while preserving most optimization opportunities; it also enables \sys's probabilistic equivalence verifier.
%

\paragraph{Expression-guided \graph generator.}
For each \lax subprogram, \sys's {\em expression-guided generator} exhaustively searches for possible \graphs equivalent to it.
A key challenge \sys must address is its significantly larger search space compared to prior superoptimization techniques.
%
For example, TASO~\cite{TASO} and PET~\cite{wang2021pet} search only for tensor programs at the kernel level, using a fixed set of pre-defined kernels, while \sys considers superoptimization across the kernel, thread block, and thread levels.
%
To efficiently navigate this significantly larger search space, \sys introduces a novel pruning technique based on {\em abstract expressions}, which greatly reduces the number of \graphs \sys must consider while providing a certain theoretical guarantee on the optimality of the discovered \graphs.
\sys further reduces the search space by focusing the search on the kernel and block levels and using a rule-based approach for the thread level. 


%
%
%

\paragraph{Probabilistic equivalence verifier.}
For a \graph discovered by \sys, verifying its functional equivalence with the input program introduces another challenge, since the input and output tensors of a program include up to many millions of elements.
A key idea behind \sys is {\em probabilistic equivalence verification}, which performs random tests over finite fields to check equivalence between \graphs.
%
While random tests typically provide limited correctness guarantees for general programs, \sys leverages a novel theoretical result showing that the restrictions imposed by the \lax fragment ensure that, for \lax programs, random tests over finite fields offer strong correctness guarantees.
Specifically, we show that a polynomial identity testing (PIT) algorithm~\cite{schwartz1980fast, zippel1979probabilistic} can be generalized to \lax programs, yielding a randomized algorithm for \lax program equivalence that can be made arbitrarily precise.
\sys uses this randomized algorithm to (probabilistically) ensure that each optimized program is equivalent to the input program.
%
%

\paragraph{\graph optimizer.} For each verified \graph, \sys's {\em \graph optimizer} maximizes its runtime performance by further considering potential tensor layouts, scheduling operator execution orders, and planning memory allocation at all of the kernel, thread block, and thread levels.
Finally, \sys returns an optimized tensor program based on the best discovered \graph for each individual \lax subprogram.

\paragraph{Evaluation results.}
We evaluate \sys on a variety of commonly used DNN benchmarks on NVIDIA A100 and H100 GPUs.
Even for DNN benchmarks that are widely used and heavily optimized by existing systems, such as the group-query attention used in LLMs~\cite{grattafiori2024llama3herdmodels}, \sys still outperforms current approaches by up to 3.3$\times$ by exploiting subtle custom kernels and optimizations missing in existing systems.
%
%

\section{Multi-Level Graph Representation}
\label{sec:hcg}

\begin{figure}
    \subfloat[Computation graph for RMSNorm and MatMul.] {
    \includegraphics[scale=0.3]{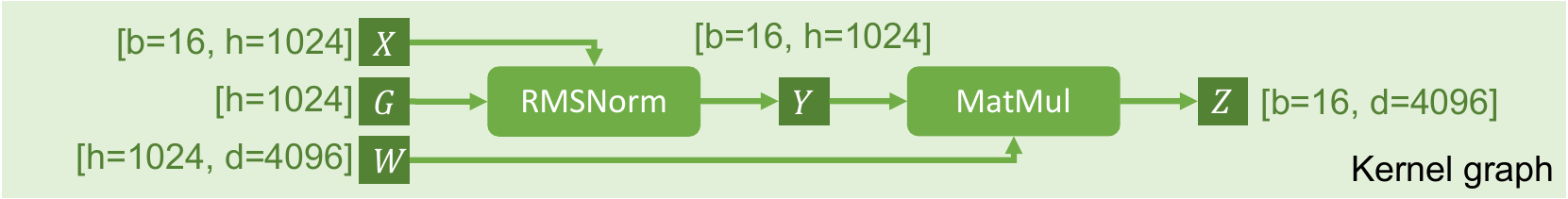}
    \label{fig:rms_norm_baseline}
    }
    \\
    \subfloat[The best \graph discovered by \sys.]{
    \includegraphics[scale=0.3]{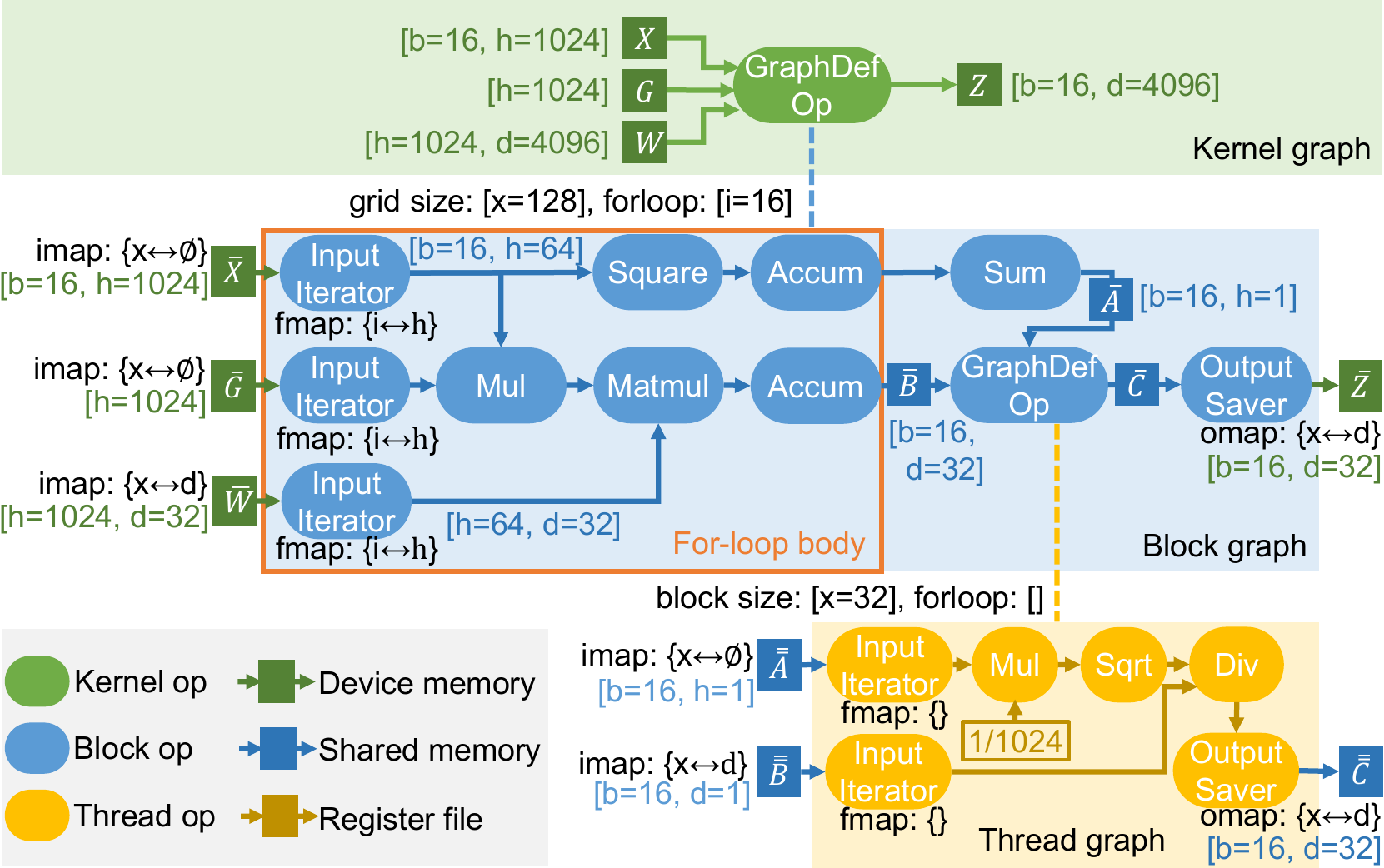}
    \label{fig:rms_norm_mlso}
    }
    \caption{\Cref{fig:rms_norm_baseline} is the computation graph for RMSNorm and MatMul. \Cref{fig:rms_norm_mlso} shows the best \graph discovered by \sys for computing RMSNorm and MatMul, which fuses the computation in a single kernel to reduce device memory access and kernel launch overhead, outperforms existing approaches by 1.9$\times$. Numbers in brackets indicate tensor shapes, and numbers in braces show the \imap, \omap, or \fmap for the corresponding operators. }
    \vspace{\captionvspace}
    \label{fig:mugraph}
\end{figure}

\sys uses a \graph to specify the execution of a tensor program on GPUs.
%
A \graph contains hierarchical graphs at multiple levels to represent computation at the kernel, block, and thread levels\footnote{For simplicity, we use the term {\em block} to refer to a thread block of a CUDA kernel and {\em thread} to refer to a single CUDA thread.}.
This section first describes the GPU hierarchy and uses \Cref{fig:mugraph} as a running example to introduce the key components of a \graph. 


\begin{figure}[t]
    \centering
    \includegraphics[width=\linewidth]{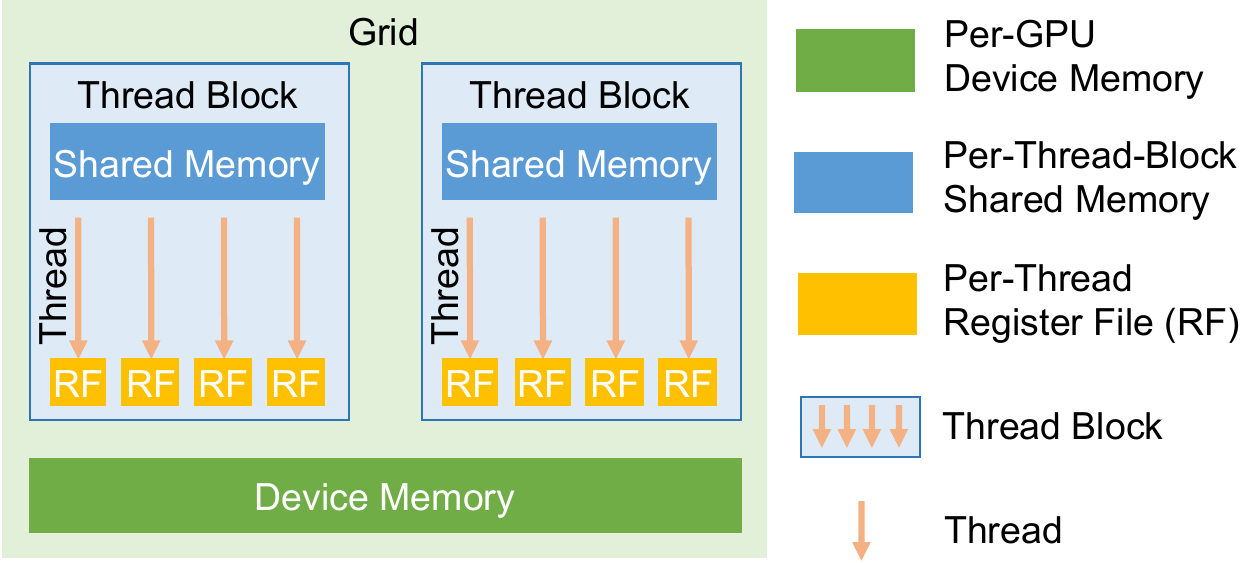}
    \caption{GPU compute and memory hierarchy.}
    \vspace{\captionvspace}
    \label{fig:gpu_hierarchy}
\end{figure}

\paragraph{GPU hierarchy.} \Cref{fig:gpu_hierarchy} shows the hierarchy of today's GPUs. Computations on GPUs are organized as {\em kernels}, each of which is a function executed simultaneously on multiple GPU cores in a single-program-multiple-data (SPMD) fashion.
A kernel includes a grid of {\em thread blocks}, each of which is executed on one GPU streaming multiprocessor and includes multiple {\em threads} to perform computation on individual data elements. 
Each thread is associated with a per-thread {\em register file}, and all threads within a thread block can access {\em shared memory} to enable collective operations.
Finally, all inputs and outputs of a kernel are stored in GPU {\em device memory}.

\paragraph{Kernel graph.}
Each tensor program corresponds to one {\em kernel graph}, where each node represents a kernel running on an entire GPU, and each edge is a tensor shared between kernels. All tensors in a kernel graph are stored in GPU device memory since different kernels cannot share data in register files or shared memory.
Each node in a kernel graph can be a {\em pre-defined} kernel operator supported by existing kernel libraries such as convolution by cuDNN~\cite{cudnn} and matrix multiplication by cuBLAS~\cite{cublas}.
%
In addition, to enable fine-grained inter-kernel optimizations such as kernel fusion, a node in a kernel graph can also be a {\em graph-defined} kernel operator, whose semantic and behavior are defined by a lower-level (i.e., block) graph.
As an example, the kernel operator in \Cref{fig:rms_norm_mlso} is a graph-defined operator specified by a block graph.

\paragraph{Block graph.} A {\em block} graph specifies computation associated with a thread block\footnote{In the CUDA programming model, a kernel's computation is defined as computations for independent thread blocks.}, where each node denotes a {\em block operator} specifying computation within a block, and each edge (blue arrows in \Cref{fig:rms_norm_mlso}) is a tensor shared between block operators.
\sys stores all intermediate tensors within a block graph in GPU {\em shared memory} for two considerations.
%
First, GPU shared memory offers much higher bandwidth than device memory, and this design allows \sys to reduce device memory access by maximally saving intermediate results in shared memory.
Second, for tensors whose sizes exceed shared memory capacity and must be stored in device memory, \sys uses these tensors to split computation into multiple block graphs, each of which only contains tensors in shared memory. This separation does not introduce additional access to device memory.

Each block graph is also associated with properties specifying its execution, which we introduce below.

\begin{figure}
    \centering
    \subfloat[$\er{imap} = \{x\leftrightarrow \er{column}\}$, $\er{fmap}=\{i\leftrightarrow \er{row}\}$]{
        \includegraphics[scale=0.42]{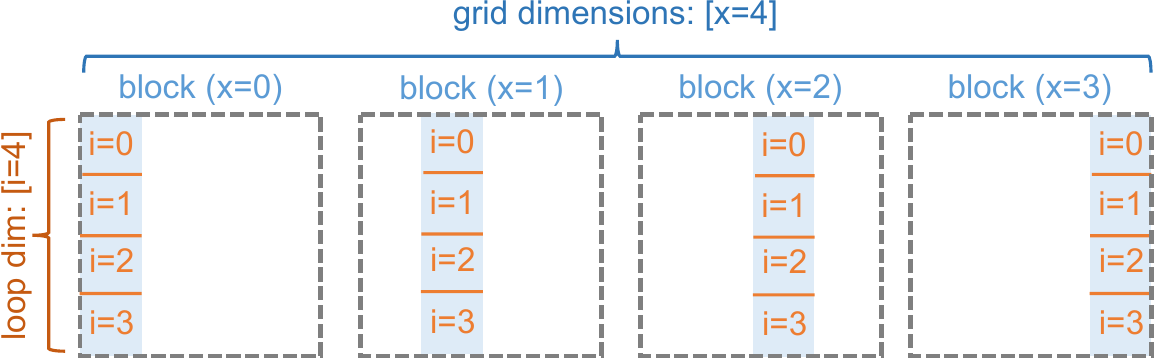}
    }
    \\
    \subfloat[$\er{imap}=\{x\leftrightarrow \phi, y\leftrightarrow \er{row}\}$, $\er{fmap}=\{i\leftrightarrow \er{column}\}$]{
        \includegraphics[scale=0.42]{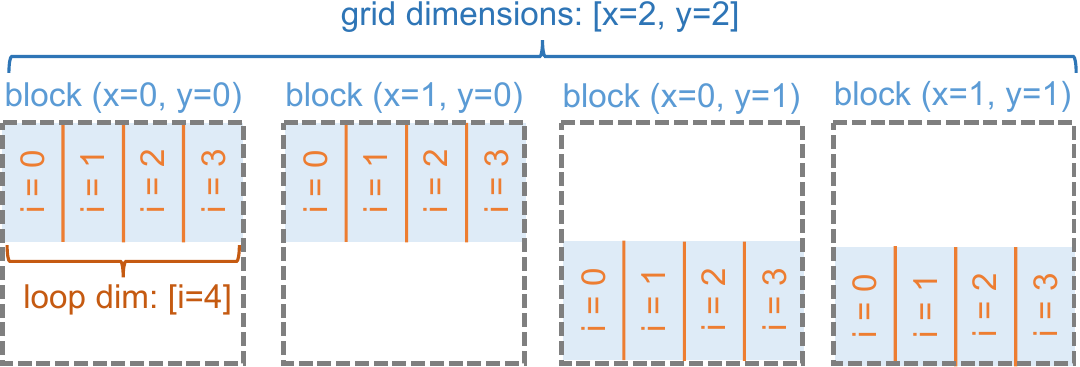}
    }
    \vspace{\captionvspace}
    \caption{Demonstrating how an input tensor is partitioned across blocks and for-loop iterations with \imap and \fmap.}
    \vspace{\captionvspace}
    \label{fig:mapping}
\end{figure}

\paragraph{Grid dimensions.} All blocks within a kernel are organized into a mesh with up to 3 dimensions, identified as $x$, $y$, and $z$. A block graph is associated with up to three {\em grid dimensions} that specify the number of blocks along the $x$, $y$, and $z$ dimensions. The block graph in \Cref{fig:rms_norm_mlso} launches $128$ blocks.
%

First, for each input tensor to a graph-defined kernel operator (e.g., $X$, $G$, and $W$ in the kernel graph in \Cref{fig:rms_norm_mlso}), the associated block graph contains an {\em \imap}, which specifies how the input tensor is partitioned into sub-tensors for individual blocks.
For each grid dimension (i.e., $x$, $y$, or $z$),
the \er{imap} maps it to either (1) a data dimension of the input tensor or (2) a special {\em replica} dimension $\phi$.
For (1), the mapped data dimension is {\em equally partitioned} across blocks along the grid dimension. For (2), the input tensor is {\em replicated} across these blocks.
For example, the block graph in \Cref{fig:rms_norm_mlso} takes three inputs---$\overline{X}$, $\overline{G}$, and $\overline{W}$---representing the input tensors to each block.
For $\overline{W}$, its $\imap=\{x\leftrightarrow d\}$ indicates that the $d$ dimension of tensor $W$ is partitioned into 128 equally sized chunks. As a result, $\overline{W}$ has shape $[h=1024, d=32]$.

Second, for each output tensor of a block graph (e.g., $\overline{Z}$ in \Cref{fig:rms_norm_mlso}), the block graph includes an {\em \omap}, which specifies how the outputs of all blocks are concatenated to construct the final output of the kernel operator.
In an \omap, each grid dimension must map to a data dimension of the output tensor, since different blocks must store disjoint tensors in device memory.
For $\overline{Z}$ with shape $[b=16, d=32]$ in \Cref{fig:rms_norm_mlso}, its $\omap =\{x\leftrightarrow d\}$ indicates that blocks with the same $x$ index are concatenated along the $d$ dimension, resulting in a tensor $Z$ with shape $[b=16, d=4096]$.

\paragraph{For-loop body.} To fit large input tensors in shared memory and to overlap data loading from device memory with computation, a block graph can include a {\em for-loop body}, which is executed multiple times to complete a kernel.
Often, the for loop in a kernel is followed by some post-processing. For example, when computing an average value, the for loop would perform the summation of $n$ values and the post-processing would divide by $n$. 
\Sys specifies the for-loop body of a block graph using {\em input iterators}, {\em for-loop accumulators}, and all operators in between, as shown in the orange box in \Cref{fig:rms_norm_mlso}). 
Each input tensor to a block graph first passes through an {\em input iterator}, which loads part of the tensor (e.g., $\overline{X}$, $\overline{G}$, and $\overline{W}$) from device memory into shared memory.
Each input iterator is associated with an {\em \fmap} to specify which part of the input tensor to load in each iteration. Formally, the \er{fmap} maps each for-loop dimension to either (1) a data dimension of the input tensor or (2) the replica dimension $\phi$. 
Similar to \imap, the tensor is equally partitioned along that dimension for (1) and replicated for (2).
\Cref{fig:mapping} shows how an input matrix is partitioned across blocks and for-loop iterations using different \imap and \fmap.

Each block graph is also associated with a {\em for-loop dimension}, which determines how many iterations the for-loop body is executed to complete the kernel. In addition, \sys uses {\em for-loop accumulators} (e.g., the two {\tt Accum} operators in \Cref{fig:rms_norm_mlso}) to accumulate intermediate results computed in each iteration (using standard accumulators, e.g., summation and max) and store the accumulated results in shared memory. Once the for-loop body is completed, \sys proceeds to execute the remaining operators outside the for-loop body directly on the accumulated results. An {\em output saver} then saves the final result from shared memory back to device memory.

\ZJ{Old version: \bf{For-loop dimensions.} To fit large input tensors in shared memory and overlap data loading with computation, a second property associated with each block graph is {\em for-loop dimensions}, which collectively specify how many times the for-loop body of the block graph is executed to complete a kernel.
\Sys defines the for-loop body of a block graph using {\em input iterators}, {\em for-loop accumulators}, and all operators in between.
Specifically, each input tensor to a block graph first passes through an {\em input iterator}, which loads part of the tensor (e.g., $\overline{X}$, $\overline{G}$, and $\overline{W}$) from device memory into shared memory.
Each input iterator is associated with an {\em \fmap} to specify which part of the input tensor to load in each iteration. Formally, the \er{fmap} maps each for-loop dimension to either (1) a data dimension of the input tensor or (2) the replica dimension $\phi$. 
Similar to \imap, the tensor is equally partitioned along that dimension for (1) and replicated for (2).
\Cref{fig:mapping} shows how an input matrix is partitioned across blocks and for-loop iterations using different \imap and \fmap.
\newline 
In addition, \sys uses {\em for-loop accumulators} (e.g., the two {\tt Accum} operators in \Cref{fig:rms_norm_mlso}) to accumulate intermediate results across iterations in shared memory.
\newline
The block graph in \Cref{fig:rms_norm_mlso} has a for-loop dimension $i=16$, indicating all operators in the for-loop body (highlighted in xxx) are executed 16 times to finish the associated graph-defined kernel operator. Operators outside of the for-loop body are performed directly on the accumulated results and are thus executed once before saving the outputs of the block graph back to device memory.}

\ZJ{To be removed: In addition, a block graph contains {\em for-loop accumulators} (e.g., the two {\tt Accum} operators in \Cref{fig:rms_norm_mlso}) to accumulate its outputs across iterations in shared memory.
Similar to an input iterator, a for-loop accumulator is also associated with an \er{fmap} to specify how outputs from different iterations are combined to produce accumulated results. 
Specifically, the \er{fmap} maps each for-loop dimension either to a data dimension, which results in concatenation of outputs along that dimension, or the replica dimension $\phi$, which results in accumulation of results in shared memory.}

\paragraph{Thread graph.} A {\em thread graph} further reduces computation scope from a block to a single thread. Similar to a block graph, each thread graph is also associated with {\em block dimensions}, which specify the organization of threads within the block, and {\em for-loop dimensions}, which define the total number of iterations to finish the defined computation. Each thread graph includes {\em input iterators}, each of which loads an input tensor (e.g., $\overline{\overline{A}}$ and $\overline{\overline{B}}$ in \Cref{fig:rms_norm_mlso}) from shared memory into register files, and {\em output savers}, each of which stores an output tensor from register files back to shared memory (e.g., $\overline{\overline{C}}$).
A thread graph is the lowest-level graph in a \graph and contains only pre-defined thread operators.

\paragraph{Tensor layout.} Each tensor in the kernel, block, or thread graph is associated with a {\em tensor layout} (omitted in \Cref{fig:mugraph} for simplicity), specifying how the tensor is linearized in memory.
Note that tensor layouts affect only the performance of a \graph and have no impact on its output correctness.

\begin{definition}[\graph Validity]
A \graph $G$ is \emph{valid} if: (1) for each kernel, block, and thread operator $o\in G$, its input and output tensors match the specification of $o$; (2) all tensors in each kernel, block, and thread graph can reside in GPU device memory, shared memory, and register file, respectively; and (3) for each block and thread graph with a for-loop body, any path from an input to an output passes through exactly one input-iterator, one for-loop accumulator, and one output saver.
\end{definition}

\paragraph{Comparison with prior work.}
Prior work separately considers algebraic~\cite{TASO, wang2021pet} or schedule transformations~\cite{halide, tvm, autohalide}, while \graphs can represent both in a uniform way.
Specifically, the grid and for-loop dimensions and their corresponding mappings (i.e., \imap, \omap, and \fmap) to tensor dimensions constitute a comprehensive search space of possible schedules for graph-defined operators. The hierarchical graphs across the kernel, block, and thread levels allow \sys to explore algebraic transformations at these levels.

\section{Case Study: RMSNorm}
\label{sec:case2}

In this section, we use root mean square layer normalization (RMSNorm)~\cite{zhang2019rootmeansquarelayer} as a case study to demonstrate the advantages of the \graph representation and \sys's superoptimization approach.
RMSNorm is a widely used normalization technique in recent large language models~\cite{grattafiori2024llama3herdmodels}. Formally, RMSNorm takes two tensors, $X$ and $G$, as inputs and normalizes their element-wise products according to the root mean square:
\begin{equation}
    Y_{ij} = \frac{X_{ij}G_j}{\textrm{RMS}(X_i)}, \textrm{RMS}(X_i) = \sqrt{\frac{1}{d}\sum_{j=1}^{d} X_{ij}^2},
\end{equation}
where $d$ is the hidden dimension size of $X$.

\begin{figure}
    \centering
    \includegraphics[width=\columnwidth]{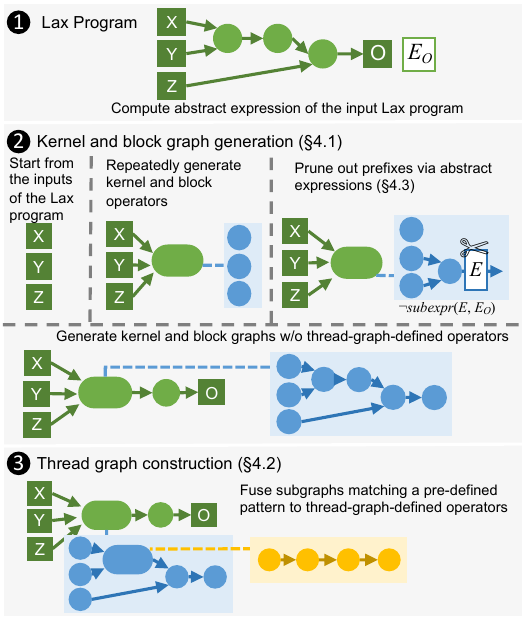}
    \vspace{-1em}
    \caption{An overview of the \graph generator.}
    \vspace{\captionvspace}
    \label{fig:generator}
\end{figure}

RMSNorm is often followed by a matrix multiplication (MatMul). \Cref{fig:rms_norm_baseline} shows the computation graph of an RMSNorm followed by a MatMul operator, where $X$ is the input tensor, and $G$ and $W$ denote two weight tensors.
Existing ML compilers generally launch two separate kernels for {RMSNorm} and MatMul computations, since both operations internally perform reductions across an input dimension, making it challenging to fuse their computations into a single kernel.
This approach requires storing intermediate results (i.e., $Y$) in device memory since different kernels cannot share data in shared memory or register files.

\Cref{fig:rms_norm_mlso} shows the best \graph automatically discovered by \sys for computing RMSNorm and MatMul in a single kernel. The computation is fused in a single graph-defined kernel operator to avoid saving intermediate results (i.e., $Y$) in device memory and reduce kernel launch overheads. 

We highlight the key differences between the \graph discovered by \sys and the original \graph.
These differences involve discovering new custom kernels and combining algebraic and schedule transformations, making it infeasible to discover the final \graph by separately considering algebraic and schedule transformations.
First, \sys reorders MatMul and the division of RMSNorm by leveraging the commutativity of matrix multiplication and element-wise division (algebraic transformation).
Second, \sys performs the accumulation in the root mean square (i.e., $A_i = \sum_j X_{ij}^2$) and the accumulation in the matrix multiplication (i.e., $B_{ik} = \sum_j X_{ij} G_j W_{jk}$) in parallel (schedule transformation), avoiding writing the accumulation results to device memory.
Next, \sys instantiates a thread graph to perform a sequence of element-wise operators while maintaining all intermediate results in register files (schedule transformation).
Finally, the best discovered \graph uses a new custom kernel to fuse the computation of RMSNorm and MatMul, reducing device memory access and kernel launch overheads. This \graph outperforms the hand-written kernels in existing systems by 1.5$\times$ and 1.9$\times$ on NVIDIA A100 and H100 GPUs respectively.

\section{Expression-Guided \graph Generator}
\label{sec:search}

\begin{algorithm}[h]
\caption{\sys's hybrid \graph generation algorithm.}
\label{algo:generate}
\footnotesize
\begin{algorithmic}[1]
\Require{A \lax program with a computation graph $\refgraph$}
\Ensure{A set of \graphs $\mathcal{S}$}
\State $E_O \gets E(\refgraph)$
\State $\mathcal{S}_0, \mathcal{S} \gets \varnothing$ \label{line:init_prefix}
\State \Call{GenerateNextKernelOperator}{$\func{Inputs}(\refgraph)$}
\ForAll{$G \in \mathcal{S}_0$} \label{line:begin_thread_graph_construct}
\State $\mathcal{S} \gets \mathcal{S} \cup \{\Call{ThreadGraphConstruction}{G}\}$\label{line:end_thread_graph_construct}
\EndFor
\Statex
\Function{GenerateNextKernelOperator}{$\kngraph$}\label{line:begin_kn}
    \State $\mathcal{S}_0 \gets \mathcal{S}_0 \cup \{\kngraph\}$
    \ForAll{kernel graph op type $t$; input set $I$}
        \If{$\func{rank}(I, t) > \func{rank}(\vi{op}.I, \vi{op}.t)$ for each $\vi{op} \in \kngraph$} \label{alg:line:kn_order}
        \If{$t$ is a pre-defined operator} \label{line:pre-defined}
          \If{$o := \Call{ConstructOp}{\kngraph, I, t}$ is valid}
              \State \Call{GenerateNextKernelOperator}{$\kngraph \cup \{o\}$}
          \EndIf
        \Else
          \Comment{$t$ is a graph-defined operator}
          \ForAll{$\vi{gridDims}$; $\vi{forloopDims}$} \label{line:pre-defined}
            \State $\tbgraph \gets \func{TBGraph}(I, \vi{gridDimd}, \vi{forloopDims})$
            \State \Call{GenerateNextBlockOperator}{$\kngraph, \tbgraph$}
          \EndFor
        \EndIf
        \EndIf
    \EndFor
\EndFunction \label{line:end_kn}
\Statex
\Function{GenerateNextBlockOperator}{$\kngraph, \tbgraph$} \label{line:begin_tb}
    \If{all shared tensors in $\tbgraph$ are consumed}
        \If{$o := \Call{ConstructOp}{\kngraph, \tbgraph.I, \tbgraph}$ is valid}
            \State \Call{GenerateNextKernelOperator}{$\kngraph \cup \{o\}$}
        \EndIf
    \EndIf
    \ForAll{block graph op type $t$; input set $I$}
        \If{$\func{rank}(I, t) > \func{rank}(\vi{op}.I, \vi{op}.t)$ for each $\vi{op} \in \tbgraph$} \label{alg:line:tb_order}
          \If{$o := \Call{ConstructOp}{\tbgraph, I, t}$ is valid}
              \State \Call{GenerateNextBlockOperator}{$\kngraph, \tbgraph \cup \{o\}$}
          \EndIf
        \EndIf
    \EndFor
\EndFunction \label{line:end_tb}
\Statex
\Function{ConstructOp}{$\ngraph, I, \vi{attrs}$} \label{line:begin_op}
    \State $E \gets \Call{ExprInfr}{E(I), \vi{attrs}}$ \Comment{Refer to Table~\ref{tab:expression}}
    \If{$\Call{Subexpr}{E, E_O}$} \Comment{Prune via abstract expressions}
    \label{line:expr_check}
    \State $S \gets \ngraph.\func{outputTensorShapeInfr}(I, \vi{attrs})$
    \Comment{Check tensor shape} \label{line:tensor_shape_check}
    \If{$S.\func{valid}$, $\ngraph.\func{mAlloc}+S.\func{size} \le \ngraph.\func{mLimit}$}
        \Comment{Check memory} \label{line:memory_check}
            \State \Return $\ngraph.\func{constructOp}(I, \vi{attrs})$
        \EndIf
    \EndIf
    \State \Return Invalid
\EndFunction \label{line:end_op}
\Statex
\Function{ThreadGraphConstruction}{$G$}
    \State $\GF{fused} \gets G$
    \While{$\exists o \in \GF{fused}$ that can be fused with a preceding operator}
        \State $\GF{fused} \gets \Call{FuseOp}{\GF{fused}, o}$
    \EndWhile
    \State \Return $\GF{fused}$
\EndFunction

\end{algorithmic}
\end{algorithm}

This section introduces the \sys \graph generator, which automatically discovers potential \graphs for an input tensor program.
To generate \graphs that capture optimizations at the kernel, block, and thread levels, \sys must explore a significantly larger search space than existing superoptimizers, which only consider optimizations at the kernel level. 
\sys employs two key techniques to address this challenge.
First, based on the observation that optimizations at the kernel and block levels are substantially more critical to performance than optimizations at the thread level---since accessing device and shared memory is orders of magnitude more expensive than accessing register file---\sys's \graph generator employs a {\em hybrid approach}: it exhaustively considers all possible graphs up to a certain size at the kernel and block levels, and uses a rule-based strategy to construct graphs at the thread level. This approach reduces the search space while retaining most performance-critical optimizations.
Second, to further prune the search space, \sys introduces a pruning technique based on an abstraction of \graphs called {\em abstract expression}, which reduces the number of \graphs \sys must consider while providing a certain theoretical guarantee on the optimality of the discovered \graphs.
We introduce the hybrid \graph generation algorithm in \S\ref{subsec:kernel_block_graph_generation} and \S\ref{subsec:thread_graph_construction}, and the expression-guided pruning techniques in \S\ref{subsec:abstract_expression}.

\subsection{Kernel and Block Graph Generation}
\label{subsec:kernel_block_graph_generation}
\sys generates kernel and block graphs incrementally and leverages several pruning techniques to reduce the search space, as shown in the second part of \Cref{fig:generator}. 
Specifically, \sys maintains a \textit{\prefix} of a valid \graph and iteratively extends it with new operators. 
For a graph $G=(V,E)$ we say that $G'=(V',E')$ is a \emph{prefix} of $G$ if it is a subgraph of $G$ such that $\forall u \in V', \forall (v, u) \in E, v \in V'$.

To generate the next operator in the kernel graph, \sys enumerates the kernel operator type $t$ and the input tensor set $I$. If $t$ represents the graph-defined operator type, \sys  generates the associated block graph that defines its kernel computation by (1)
enumerating the grid and for-loop dimensions (introduced in \S\ref{sec:hcg}), which enables \sys to calculate the input tensor shapes of the block graph; and (2) performing a nested generation procedure similar to that used at the kernel level but without considering graph-defined operators. \Cref{line:begin_kn}-\ref{line:end_kn} and \cref{line:begin_tb}-\ref{line:end_tb} in \Cref{algo:generate} show how \sys generates kernel and block operators, respectively.
%
%
%
\sys checks tensor shape (\cref{line:tensor_shape_check}) and memory usage (\cref{line:memory_check}) before adding an operator, ensuring a valid \prefix.

To ensure that identical \graphs are generated only once, \sys defines the \textit{canonical form} of \graphs. 
Given a \graph $G$ with its operators in topological order $o_1, \ldots, o_n$, the \textit{index} of the $j$-th output of $o_i$ is defined as a tuple $(i, j)$. Each operator $o_i$ in $G$ is assigned a \textit{rank} $(\vi{input}_i, \vi{type}_i)$, where $\vi{input}_i$ is the list of input tensor indices of $o_i$, and $\vi{type}_i$ is the operator type. 
A \graph is in canonical form if its operators are ordered in increasing rank. 
\sys generates only \graphs in canonical form by requiring that operators be added in increasing order of rank (line \ref{alg:line:kn_order} and \ref{alg:line:tb_order}). This approach does not prune out any valid solutions, since each \graph can be transformed to canonical form by reordering the operators.

In addition, \sys utilizes the \textit{abstract expression} technique to prune out prefixes that do not satisfy certain constraints, which will be introduced in \S\ref{subsec:abstract_expression}.

\subsection{Thread Graph Construction}
\label{subsec:thread_graph_construction}

While a similar nested generation strategy can be applied to thread graphs, \sys instead constructs them using a transformation-based approach (see the third panel of \Cref{fig:generator} and lines \ref{line:begin_thread_graph_construct}–\ref{line:end_thread_graph_construct} in \Cref{algo:generate}) to reduce the search space. 
%
\sys applies operator fusion when constructing thread graphs, which reduces access to shared memory by reusing tensors in register file whenever possible. 
For example, \sys fuses the three element-wise operators ({\tt Mul}, {\tt Sqrt}, and {\tt Div}) in \Cref{fig:rms_norm_mlso} into a thread graph, avoiding saving intermediate results to shared memory and keeping the entire computation of these operators in register file.
While our current implementation focuses on operator fusion, additional rule-based transformations can be used to construct thread graphs.

\begin{table}
    \centering
    \caption{Operators supported by \sys. The second column shows the graph levels supporting each operator (K, B and T denote kernel, block, and thread graphs, respectively). The last column defines the abstract expressions for the outputs of each operator, where $E$ maps tensors to their abstract expressions.}
    \vspace{\captionvspace}
    \label{tab:expression}
    \resizebox{\columnwidth}{!}
    {%
    \begin{threeparttable}
    \begin{tabular}{l|l|l}
    \toprule
    {\bf \graph} & {\bf Graph} &  {\bf Abstract Expression of Output Tensor } \\
    {\bf Operator} & {\bf Level} &  {\bf } \\
    \midrule
    InIter & B & $\expr(\mathrm{InIter}(X)) = \expr(X)$ \\
    OutSaver  & B & $\expr(\mathrm{OutSaver}(X)) = \expr(X)$ \\
    Matmul & K, B, T & $\expr(\mathrm{Matmul}(X, Y)) = \ered(k, \emul(\expr(X), \expr(Y)))$\tnote{1} \\
    Sum & K, B, T & $\expr(\mathrm{Sum}(d_r, k_r, X)) = \ered(k_r, \expr(X))$\tnote{2} \\
    EwAdd & K, B, T & $\expr(\mathrm{EwAdd}(X, Y)) = \eadd(\expr(X), \expr(Y))$ \\
    EwMul & K, B, T & $\expr(\mathrm{EwMul}(X, Y)) = \emul(\expr(X), \expr(Y))$ \\
    EwDiv & K, B, T & $\expr(\mathrm{EwDiv}(X, Y)) = \ediv(\expr(X), \expr(Y))$ \\
    EwExp & K, B, T & $\expr(\mathrm{EwExp}(X)) = \eexp(\expr(X))$ \\
    Repeat & K, B & $\expr(\mathrm{Repeat}(X)) = \expr(X)$ \\
    Reshape & K, B & $\expr(\mathrm{Reshape}(X)) = \expr(X)$ \\
    Sqr & K, B & $\expr(\mathrm{Sqr}(X)) = \emul(\expr(X), \expr(X))$ \\
    Sqrt & K, B & $\expr(\mathrm{Sqrt}(X)) = \esqrt(\expr(X))$ \\
    SiLU & K, B & $\expr(\mathrm{SiLU}(X)) = \esilu(\expr(X))$ \\
    Accum & B & $\expr(\mathrm{Accum}(X, m, i)) = \ered(i, \expr(X))$ if $m = \phi$ else $\expr(X)$ \tnote{3} \\
    \bottomrule
    \end{tabular}
    \begin{tablenotes}
    \item[1] $k$ means the size of the last dimension of $A$, i.e., the reduction dimension. Matmul is performed on the inner most two dimensions and leading dimensions are batched.
    \item[2] Sum along the dimension $d_r$ for every $k_r$ elements.
    \item[3] Accumulate the results of $i$ for-loop iterations along fmap $m$.
    \end{tablenotes}
    \end{threeparttable}
    }
    \vspace{\captionvspace}
\end{table}

\subsection{Pruning via Abstract Expressions}
\label{subsec:abstract_expression}



When searching the space of possible \graphs, we aim to avoid \graph prefixes whose intermediate results cannot contribute to the desired computation.
For example, for the input program $X \cdot Z + Y \cdot Z$,
we can prune a prefix that computes $X \cdot Y$, but we should not prune one that computes $X + Y$, as $(X+Y) \cdot Z$ is equivalent to the input program.
However, how can we determine whether a prefix can contribute to a desired computation while searching for that computation?
Below, we develop a pruning technique driven by this intuition that circumvents the ``chicken and egg'' problem via \emph{abstraction}.
We first present the abstraction---\emph{abstract expressions}---and then explain how it is used for pruning.
Finally, we offer a theoretical guarantee that, under certain conditions, this pruning does not exclude the optimal \graph.

\begin{figure}
    \centering
    \includegraphics[width=0.9\columnwidth]{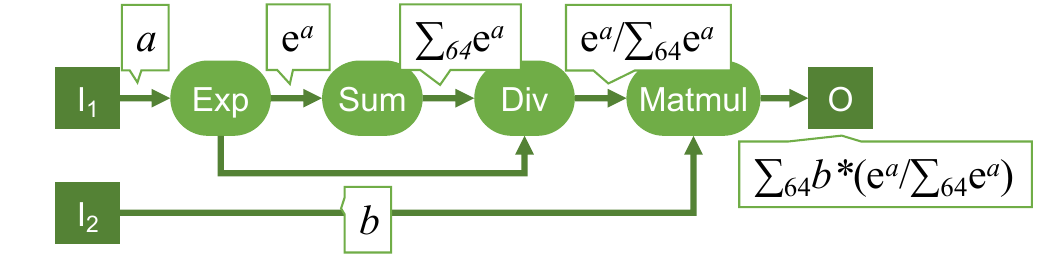}
    \vspace{\captionvspace}
    \caption{Illustration of abstract expressions. The abstract expressions of tensors are annotated on edges. A human-friendly notation is used here: $\mathrm{e}^a$ denotes $\eexp(a)$, $\sum_k a$ denotes $\ered(k, a)$, $a/b$ denotes $\ediv(a, b)$, and $a*b$ denotes $\emul(a, b)$. The tensors $I_1$, $I_2$ and $O$ are all $64\times64$ matrices.}
    \vspace{\captionvspace}
    \label{fig:abstract_expression}
\end{figure}

\paragraph{Abstract expressions.}
Recall that an edge in a \graph corresponds to a tensor-valued function of the input tensors. Intuitively, abstract expressions abstract these functions by ignoring the differences between elements of the same input tensor.
Formally, abstract expressions are first-order logic terms over the theory of integers and uninterpreted functions.
In a \graph, the abstract expression of each edge, denoted by $\expr(\cdot)$, is defined in \Cref{tab:expression}.
When computing a \graph's abstract expression, all graph-defined operators are ``inlined''.
Specifically, the expressions computed for a graph-defined operator's inputs are passed into its lower-level graph, and the resulting output expressions of that lower-level graph become the output expressions of the graph-defined operator.
\Cref{fig:abstract_expression} shows the abstract expressions for a subgraph of attention.

While abstract expressions capture some information about the function computed at each edge, they also abstract away many details. For example, if $X$ is a $k \times k$ matrix, summing over the rows and summing over the columns both yield the same abstract expression---$\ered(k,\expr(X))$. But keeping $k$ as part of the abstract expression is crucial for effective pruning.


\paragraph{Abstract subexpression and pruning.}
We use abstract expressions to prune the search space of \graphs by formalizing two relations over abstract expressions: equivalence and abstract subexpression. Specifically, we prune any \graph prefix whose abstract expression is not a subexpression of some abstract expression equivalent to that of the input program.
%
We formalize abstract expressions as uninterpreted functions in first-order logic over the theory of integer arithmetic and uninterpreted functions,
and use an SMT solver to reason about them based on two sets of axioms in \Cref{tab:axioms}: $\Aeq$ and $\Asub$.

First, $\Aeq$ axiomatizes equivalence between abstract expressions.
As will become clear below, these axioms need not be sound---it is not required that \graphs with equivalent abstract expressions are functionally equivalent, since non-equivalent \graphs can have the same abstract expression.
Second, $\Asub$ axiomatizes the subexpression relation between abstract expressions. A key property of $\Asub$ is that whenever a \graph $G_1$ is a prefix of $G_2$---meaning $G_2$ can be constructed by extending $G_1$ with additional operators---$\expr(G_1)$ is an abstract subexpression of $\expr(G_2)$; formally, $\Asub \models \subexpr(\expr(G_1), \expr(G_2))$, where $\models$ denotes entailment modulo the theory of integer arithmetic and uninterpreted functions.

During the search, \Cref{algo:generate} first computes the abstract expression of the input \lax program, denoted $E_O$, and prunes any \graph prefix $G$ if $\Aeq \cup \Asub \not\models \subexpr(\expr(G),E_O)$. That is, a graph is pruned if its abstract expression is not a subexpression of $E_O$. This check is performed using an SMT solver (Z3~\cite{z3}).
As an optimization, the results of these checks are cached and reused, since \sys may encounter multiple \graphs with identical abstract expressions during the search.

\begin{table}[h]
    \centering
    \caption{Axiomatization of abstract expressions used for pruning. \sys checks whether an abstract expression $E_1$ is a subexpression of $E_2$ by querying an SMT solver to check if $\subexpr(E_1, E_2)$ is entailed by these axioms. All variables in these axioms are universally quantified.}
    \vspace{\captionvspace}
    \label{tab:axioms}
    \resizebox{\columnwidth}{!}{
    \begin{tabular}{l|l}
    \toprule
    {\bf Abstract Expression Property} & {\bf Comment} \\
    \midrule
    \multicolumn{2}{l}{\it Equivalence Axioms $\Aeq$} \\
    \midrule
    $\forall x,y.\ \eadd(x, y) = \eadd(y, x)$ & commutativity \\
    $\forall x,y.\ \emul(x, y) = \emul(y, x)$ & commutativity \\
    $\forall x,y,z.\ \eadd(x, \eadd(y, z)) = \eadd(\eadd(x, y), z)$ & associativity \\
    $\forall x,y,z.\ \emul(x, \emul(y, z)) = \emul(\emul(x, y), z)$ & associativity \\
    $\forall x,y,z.\ \eadd(\emul(x, z), \emul(y, z)) = \emul(\eadd(x, y), z)$ & distributivity \\
    $\forall x,y,z.\ \eadd(\ediv(x, z), \ediv(y, z)) = \ediv(\eadd(x, y), z)$ & associativity \\
    $\forall x,y,z.\ \emul(x, \ediv(y, z)) = \ediv(\emul(x, y), z)$ & associativity \\
    $\forall x,y,z.\ \ediv(\ediv(x, y), z) = \ediv(x, \emul(y, z))$ & associativity \\
    $\forall x.\ x = \ered(1, x)$ & identity reduction \\
    $\forall x,i,j.\ \ered(i, \ered(j, x)) = \ered(i*j, x)$ & associativity \\
    $\forall x,y,i.\ \ered(i, \eadd(x, y)) = \eadd(\ered(i, x), \ered(i, y))$ & associativity \\
    $\forall x,y,i.\ \ered(i, \emul(x, y)) = \emul(\ered(i, x), y) $ & distributivity \\
    $\forall x,y,i.\ \ered(i, \ediv(x, y)) = \ediv(\ered(i, x), y) $ & distributivity \\
    $\forall x,y.\ \emul(\eexp(x), \eexp(y)) = \eexp(\eadd(x, y))$ & distributivity \\
    $\forall x,y.\ \emul(\esqrt(x), \esqrt(y)) = \esqrt(\emul(x, y))$ & distributivity \\
    
    \midrule
    \multicolumn{2}{l}{\it Subexpression Axioms $\Asub$} \\
    \midrule
    $\forall x,y.\ \subexpr(x, \eadd(x, y))$ & \\
    $\forall x,y.\ \subexpr(x, \emul(x, y))$ & \\
    $\forall x,y.\ \subexpr(x, \ediv(x, y))$ & \\
    $\forall x,y.\ \subexpr(y, \ediv(x, y))$ & \\
    $\forall x.\ \subexpr(x, \eexp(x))$ & \\
    $\forall x.\ \subexpr(x, \esqrt(x))$ & \\
    $\forall x.\ \subexpr(x, \esilu(x))$ & \\
    $\forall x,i.\ \subexpr(x, \ered(i, x))$ & \\
    $\forall x.\ \subexpr(x, x)$ & reflexivity \\
    $\forall x,y,z.\ \subexpr(x,y) \wedge \subexpr(y,z) \to \subexpr(x, z)$ & transitivity \\
    \bottomrule
    \end{tabular}
    }
    \vspace{\captionvspace}
\end{table}

\paragraph{Theoretical guarantee and the pruning-optimality tradeoff.}
Intuitively, our pruning would keep any prefix that can lead to a \graph whose abstract expression is equivalent (according to $\Aeq$) to that of the input \lax program. Formally:

\begin{theorem}[Pruning via Abstract Expressions]
\label{thm:pruning}
For an input \graph $G_0$, and a \graph $G$ equivalent to $G_0$,
if $\Aeq \models E(G_0) = E(G)$ then $G$ will be generated by \Cref{algo:generate}.
\end{theorem}

\iffinal
\begin{proof}
By \Cref{tab:expression,tab:axioms}, we show that for any operator $\er{op}$, if $Y = \er{op}(X_1,\ldots,X_n)$, then 
$\Asub \models \subexpr(\expr(X_i), \expr(Y))$ for $1 \leq i \leq n$.
That is, the abstract expression of each input to $\er{op}$ is always a subexpression of $\er{op}$'s output.
Given that $\Asub$ includes reflexivity and transitivity axioms,
it follows that for any $G'$ that is a prefix of $G$, $\Asub \models \subexpr(\expr(G'),\expr(G))$.
Together with the assumption that $\Aeq \models E(G_0) = E(G)$,
we have $\Aeq \cup \Asub \models \subexpr(\expr(G'),\expr(G_0))$.
Thus, no prefix of $G$ will be pruned, and \sys will generate $G$.
\end{proof}
\fi 

The theorem highlights the role of abstract expressions in solving the ``chicken and egg'' problem outlined above. To decide if a prefix \graph is useful, we reason about whether it is a prefix of a useful computation \emph{in the abstract}. The choice of abstraction and the axioms $\Aeq$ represents a tradeoff between optimality and pruning.
As \Cref{thm:pruning} shows, we are only guaranteed to find the optimal \graph whose abstract expression is equivalent to that of the input program under $\Aeq$. Stronger axioms expand the set of \graphs covered by the theorem but reduce pruning effectiveness, since more prefixes would pass the subexpression test. In particular, note that $\Aeq$ does not include cancellation rules (e.g., $\ediv(\emul(x,y),y) = y$). As a result, \sys may miss some equivalent \graphs.
However, including such axioms would make everything a subexpression of everything, therefore nulling desired pruning.
As our evaluation shows, the chosen $\Aeq$ yields a good balance between pruning and optimality.




\section{Probabilistic Equivalence Verifier}
\label{sec:verify}


\sys's {\em probabilistic equivalence verifier} checks if a candidate \graph is equivalent to the desired \lax program.
The key idea is to evaluate both on {\em random inputs} in two finite fields.
Using finite fields instead of floating point numbers not only avoids floating point errors but also provides a strong theoretical guarantee: the probability of accepting a non-equivalent \graph can be made arbitrarily low.

For general programs, random testing can hardly provide any correctness guarantee.
%
However, we show that for \lax programs (formally defined below), random testing offers a probabilistic correctness guarantee, and repeated tests can reduce the error probability to an arbitrarily small threshold.

Prior work~\cite{wang2021pet} has applied a similar technique to check equivalence between tensor programs that contain only linear operators (e.g., matrix multiplication, convolution).
We develop a random testing technique that also supports division and exponentiation, which are needed for many DNN optimizations (e.g., the RMSNorm example in \S\ref{sec:case2}).

%
%
\sys verifies equivalence between \lax \graphs (\underline{l}inear, division, and \underline{a}n e\underline{x}ponentiation) defined below. We introduce the main theoretical results in \S\ref{subsec:theory} and present \sys's verification methodology in \S\ref{subsec:random_tests}.

\begin{definition}[\lax \graph]
A \graph $G$ is a \lax \graph if (1) $G$ contains only multi-linear operators\footnote{Operator \er{op} with $n$ inputs is multi-linear if \er{op} is linear to all inputs $I_k$: \newline (1) $\forall X, Y. \er{op}(I_1,...,I_{k-1},X,I_{k+1},...,I_n)+\er{op}(I_1,...,I_{k-1},Y,I_{k+1},...,I_n)=\er{op}(I_1,...,I_{k-1},X+Y,I_{k+1},...,I_n)$, and \newline(2) $\alpha \cdot \er{op}(I_1,...,I_{k-1},X,I_{k+1},...,I_n) = \er{op}(I_1,...,I_{k-1},\alpha\cdot X,I_{k+1},...,I_n).$}, division, and exponentiation, and (2) every path from an input to an output in $G$ includes at most one exponentiation.
\end{definition}
%

\subsection{Theoretical Foundations}
\label{subsec:theory}
Without loss of generality, we assume a \lax \graph $G$ takes $n$ input tensors and produces one output tensor.
Our theoretical results directly generalize to \lax \graph with multiple outputs.
Since each \lax \graph includes linear operators, divisions, and at most one exponentiation along each path,
%
the computation for each entry of the output tensor can be expressed in the following form (by using standard identities such as $\frac{\frac{a}{b}}{\frac{c}{d}} = \frac{ad}{bc}$, $\frac{a}{b} + \frac{c}{d} = \frac{ad+bc}{bd}$, $e^{x}e^{y}=e^{x+y}$):
\begin{equation}
\label{eq:lax}
\frac{\sum_{i=1}^k f_i \exp(g_i/h_i)}{\sum_{i=1}^{k'} f'_i \exp(g'_i/h'_i)}
\end{equation}
where $f_i$, $g_i$, $h_i$, $f_j'$, $g_j'$ and $h_j'$ ($1\leq i \leq k$, $1 \leq j \leq k'$) are polynomials over the entries of the input tensors.

The main theoretical result that underpins our randomized equivalence verification is the following theorem, which extends polynomial identity testing (PIT)~\cite{schwartz1980fast, zippel1979probabilistic} on finite fields to \lax \graphs. Note that the difference of two \lax \graphs is also of the form of \Cref{eq:lax}. Therefore, identity testing of two \lax \graphs reduces to testing if an expression of that form is zero.
Due to the presence of exponentiation, we use two finite fields instead of one.\footnote{We use two primes $p$ and $q$ for polynomial identity testing \cite{schwartz1980fast, zippel1979probabilistic} outside and inside the exponents, respectively. The condition $q$ divides $p - 1$ is to ensure the existence of $q$-th roots of unity in $\mathbb{Z}_p$.}

\begin{theorem}
\label{thm:zero_test}
Let $P$ be a function of the form described in \Cref{eq:lax},
where
$f_i, g_i, h_i, f_i', g_i', h_i'$ are non-zero polynomials of degree at most $d$ with integer coefficients between $[-w, w]$. Let $p, q$ be primes such that $q \mid p-1$ and $q > 2w$. Let $\mathcal{G}$ be the set of $q$-th roots of unity in $\mathbb{Z}_p$. If $P$ is not a zero function, then \cite{identitytesting}
\[
\Pr_{(\vec u, \vec v, \omega) \gets \mathbb{Z}_p^N\times \mathbb{Z}_q^N \times \mathcal{G}}\left[\frac{\sum_{i=1}^k f_i(\vec u) \omega^{g_i(\vec v)/h_i(\vec v)}}{\sum_{i=1}^{k'} f'_i(\vec u) \omega^{g'_i(\vec v)/h'_i(\vec v)}}\right] \le 8dk^4/q + q^{-1/k^2}.
\]
\end{theorem}

\iffinal
\fi 

\subsection{Random Tests over Finite Fields}
\label{subsec:random_tests}

\begin{table}
    \centering
    \caption{Arithmetic operations for random testing. \sys selects two prime numbers $p$ and $q$ such that $q$ divides $p - 1$. $x_p$ and $x_q$ are values from the finite fields $\mathbb{Z}_p$ and $\mathbb{Z}_q$, respectively. The notation $x^{-1}$ and $\sqrt{x}$ represents the multiplicative inverse and square root of $x$ in the corresponding finite field. Specifically,  $xx^{-1} \bmod p = 1$ and $\sqrt{x}\sqrt{x} \bmod p = x$.}
    \vspace{\captionvspace}
    \label{tab:fingerprint}
    \resizebox{\columnwidth}{!}    
    {%
    \footnotesize
    \begin{tabular}{c|c|c|l}
    \toprule
    {\bf Opt.} & {\bf Opd. 1} & {\bf Opd. 2} & {\bf Output} \\
    \midrule
    Add. & $(x_p, x_q) $ & $(y_p, y_q)$ & $\big((x_p + y_p) \bmod{p}, (x_q + y_q) \bmod{q}$\big) \\
    Sub. & $(x_p, x_q) $ & $(y_p, y_q)$ & $\big((x_p - y_p) \bmod{p}, (x_q - y_q) \bmod{q}$\big) \\
    Mul. & $(x_p, x_q) $ & $(y_p, y_q)$ & $\big(x_py_p \bmod{p}, x_qy_q\bmod{q}\big)$ \\
    Div. & $(x_p, x_q) $ & $(y_p, y_q)$ & $\big(x_py_p^{-1} \bmod{p}, x_qy_q^{-1}\bmod{q}\big)$ \\
    Exp. & $(x_p, x_q) $ & $ - $ & $\big(\omega^{x_q} \bmod{p}, -\big)$ \\
    Sqrt. & $(x_p, x_q)$ & $ - $ & $(\sqrt{x_p}, \sqrt{x_q})$ \\
    \bottomrule
    \end{tabular}
    }
\end{table}

\sys leverages \Cref{thm:zero_test} to probabilistically verify the equivalence of two \graphs by performing random testing over the finite fields $\mathbb{Z}_p$ and $\mathbb{Z}_q$ as defined in \Cref{thm:zero_test}.
To check the equivalence of two \graphs, \sys first generates input tensors, with each entry uniformly sampled from $\mathbb{Z}_p \times \mathbb{Z}_q$.
\sys also samples $\omega$ uniformly from the set of $q$-roots of unity in $\mathbb{Z}_p$, which is used for exponentiation.
\sys then evaluates the two \graphs on these inputs using the operations defined in \Cref{tab:fingerprint}.
As explained in \S\ref{subsec:theory}, $\mathbb{Z}_p$ and $\mathbb{Z}_q$ are used for computations outside and inside the exponent, respectively.
All operations except exponentiation are implemented via modular arithmetic in $\mathbb{Z}_p$ and $\mathbb{Z}_q$ independently.
For exponentiation, \sys uses the value $x_q$ from $\mathbb{Z}_q$ and computes $\omega^{x_q} \bmod p$ to obtain a result in $\mathbb{Z}_p$.

Note that in a \lax \graph, exponentiation is performed at most once along each path.
Finally, \sys checks whether the two \graphs produce identical outputs.
This process is repeated multiple times, and the two \graphs are considered equivalent if they pass all random tests.
The following theorem, which follows from \Cref{thm:zero_test}, shows that this process can yield an arbitrarily low error rate.

\begin{theorem}
\label{thm:prob}
Equivalent \graphs always pass \graph verification. 
For two non-equivalent \graphs and a given probability threshold $0<\delta\leq 1$, the \graphs pass all $\Omega(\frac{k^2}{\ln q}\cdot \ln\frac{1}{\delta})$ random tests with probability at most $\delta$.
\end{theorem}

\paragraph{Numerical stability.} While the theorem bridges finite fields and real-number computations, discrepancies can arise between real-number computations and floating-point operations, particularly involving overflow or underflow due to large intermediate values. \sys employs floating-point tests to filter out \graphs with significant numerical errors.

\section{\graph Optimizer}
\label{sec:optimizer}

For each verified \graph, \sys's {\em \graph optimizer} maximizes its performance by further performing {\em layout optimization}, {\em operator scheduling}, and {\em memory planning}, as shown in \Cref{fig:overview}. 
\sys defers these \graph optimizations until after verification for two reasons.
First, these optimizations {\em do not} affect the correctness of the generated \graphs; omitting them when generating \graphs reduces the search space \sys must consider, since \graphs with the same graph topology but different choices of tensor layouts, operator orders, or memory allocation plans are considered identical by the \graph generator.
Second, applying these optimizations after verification also reduces the search space for these optimizations, since the \graph optimizer only needs to optimize \graphs that are functionally equivalent to the input.

\paragraph{Tensor layouts.} The \graph optimizer explores possible data layouts for all intermediate tensors at the kernel, block, and thread levels and chooses the best combinations to maximize performance. We formulate layout selection as a constrained optimization problem and solve it optimally using an integer linear programming (ILP) algorithm.
Specifically, for each tensor $t$ and each possible layout $l$ for $t$, we introduce a boolean variable $B_{t,l}$ to indicate whether tensor $t$ uses layout $l$. 
Operators at the kernel, block, and thread levels may impose various constraints on tensor layouts.
For example, to use kernels from the cuBLAS library~\cite{cublas} for matrix multiplication, the innermost dimension of the two input tensors must be among the last two dimensions. These restrictions are converted into a series of linear constraints on $B_{t, l}$. Different tensor layouts may lead to varying performance. For example, some input tensor layouts support bulk copies from device to shared memory, while others do not. \sys introduces a cost function to model the performance of each operator under different layout choices.
\sys uses an off-the-shelf ILP solver (i.e., Z3~\cite{z3}) to find an optimal layout strategy that satisfies all layout constraints while minimizing cost.

\paragraph{Operator scheduling.} In a \graph, there are multiple topological orders to execute operators, and different orders may yield different performance. 
For a given input \graph, the \graph optimizer identifies an efficient strategy to schedule operators by minimizing thread-level synchronization within each thread block (i.e., {\tt \_\_syncthreads()} in CUDA). 
To achieve this goal, \sys labels each node with a {\em depth}, defined as the length of the longest path from any input operator to that node. \sys uses a dynamic programming algorithm to compute the depth of each node and schedules all operators in ascending order of their depths.
This approach minimizes the number of thread-level synchronizations required in the generated CUDA kernel, as \sys only needs to insert synchronization points between operators with different depths.

\paragraph{Memory planning.} A third class of post-verification optimizations is memory planning, which determines memory offsets for all intermediate tensors at the kernel, block, and thread levels. \sys formulates memory planning as a {\em dynamic storage allocation} problem and exhaustively enumerates all possible allocation plans to discover an optimal strategy.

\section{Implementation}
\label{sec:impl}

\begin{table}
    \centering
    \caption{DNN benchmarks used in our evaluation.}
    \vspace{\captionvspace}
    \footnotesize
    \label{tab:benchmarks}
    \begin{tabular}{l|l|l}
    \toprule
    {\bf Name} & {\bf Description} & {\bf Base Architecture} \\
    \midrule
    GQA & Group-query attention & \llama-3-70B~\cite{grattafiori2024llama3herdmodels}\\
    QKNorm & QK normalization with attention & Chameleon-7B~\cite{chameleonteam2024chameleon} \\
    RMSNorm & RMS normalization with linear & \llama-2-7B~\cite{llama2}\\
    LoRA & Low-rank adaptation & GPT-3-7B-LoRA~\cite{llama-7b-lora} \\
    GatedMLP & Gated multi-layer perceptron & Falcon-7B~\cite{falcon40b}\\
    nTrans & Normalized Tarnsformer & nGPT-1B~\cite{loshchilov2024ngpt}\\

    \bottomrule
    \end{tabular}
    \vspace{\captionvspace}
    \vspace{\captionvspace}
\end{table}

\sys is implemented in 30K lines of code in C++, CUDA, and Python. Kernel operators are implemented with the cuDNN and cuBLAS libraries~\cite{cudnn, cublas}, and block and thread operators are implemented using cuTLASS~\cite{cutlass} and CUDA PTX.
For each input tensor program, \sys automatically generates and verifies potential \graphs. 
For each verified \graph, \sys produces CUDA source code for all custom kernels of the \graph and compiles the code into binary using the CUDA compiler.
This approach enables just-in-time (JIT) compilation and deployment for general tensor programs, and the generated kernels can be directly integrated into a PyTorch program with a few lines of code changes. 
\sys's SMT and ILP solvers are implemented using Z3 4.12.6~\cite{z3}. 

Our implementation supports the operators listed in \Cref{tab:expression}.
%
\sys can be extended to include new operators, such as variants of convolution or matrix multiplication, at the kernel, block, and/or thread levels.
To support a new linear operator, \sys requires (1) a float-pointing implementation of the operator at the kernel, block, and/or thread levels, which is used by the \graph optimizer to generate CUDA kernels; (2) an implementation of the operator over modular arithmetic (see \S\ref{sec:verify}); and (3) an extension to the abstract expression axioms $\Aeq$ and $\Asub$ for the operator (see \S\ref{subsec:abstract_expression}).
%

To utilize \Cref{thm:zero_test,thm:prob}, random tests should be performed with sufficiently large prime numbers $p$ and $q$ and iterated multiple times. Our current implementation uses the largest values of $p$ and $q$ whose product fits in 16-bit integers (i.e., $p=227, q=113$) to run these random tests on GPUs. We leverage \sys's GPU optimizations--such as keeping intermediate results in shared memory--to accelerate the search procedure.
We also perform a single random test without iterating it and compare all elements of the output tensors.
We note that this equivalence verification procedure does not introduce false negatives.
While it could, in theory, introduce false positives, we have not observed any in practice.
For these reasons, we consider this procedure sufficient for the search process and plan to add a final verification step that provides the theoretical guarantees only for the best \graph at the end of the optimization process.

\paragraph{Equivalence verification for non-\lax programs.} While \sys can generate \graphs for arbitrary tensor programs, the probabilistic equivalence verifier is limited to \lax programs and does not support certain DNN operators such as ReLU~\cite{relu}. 
As an alternative, we have developed a solver-based verifier for arbitrary tensor programs. The verifier relies on user-provided mathematical properties of individual operators (e.g., linearity, associativity, commutativity, and distributivity) defined in first-order logic and uses these properties to verify equivalence using an automated theorem prover. Compared to the probabilistic equivalence verifier, the solver-based verifier supports more general programs, while requiring additional manual effort to specify the properties of each new operator. A detailed discussion of the solver-based verifier is beyond the scope of this paper.

\section{Evaluation}
\label{sec:eval}

\subsection{Experimental Setup}
\label{subsec:eval_setup}

\begin{figure*}
    \centering
    \includegraphics[width=\textwidth]{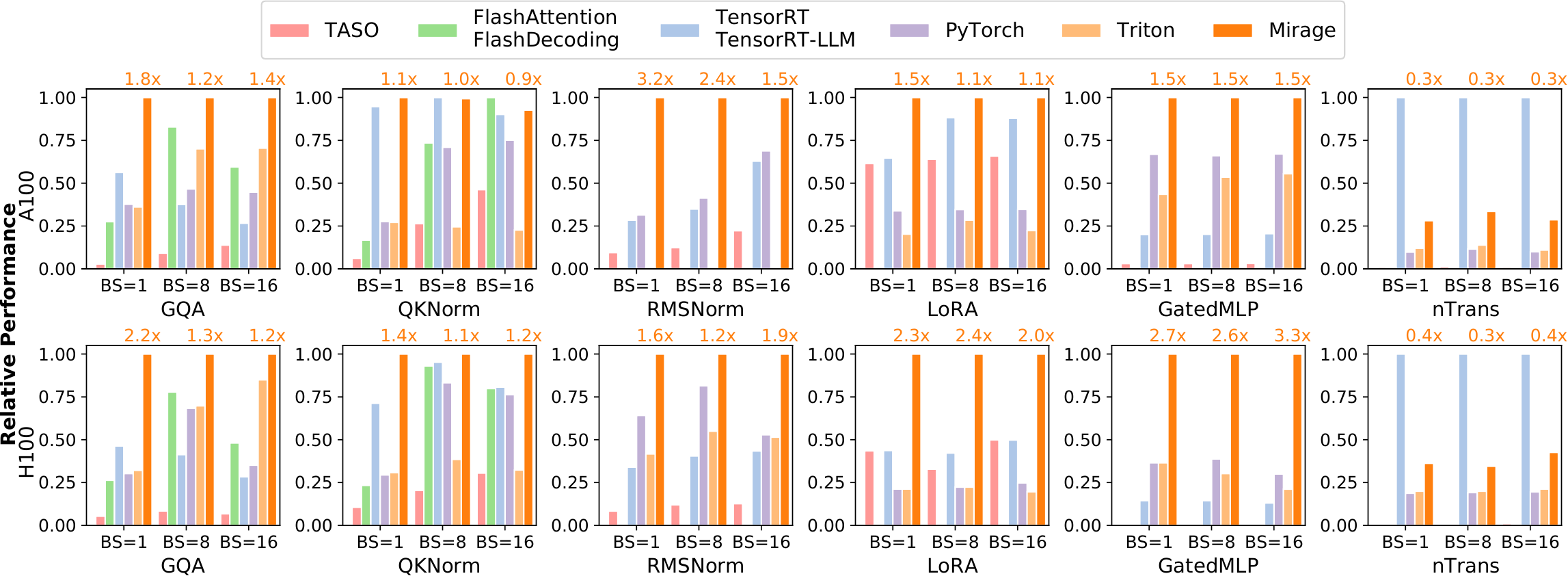}
    \vspace{-1em}
    \caption{Comparing \sys with existing systems for 6 benchmarks on an A100 and an H100 GPU. The performance of all systems are normalized by \sys (higher is better). Numbers above the \sys bars show the speedup over the best baselines.}
    \vspace{\captionvspace}
    \label{fig:benchmarks}
\end{figure*}

Since \sys is a superoptimizer for \lax programs, we focus our evaluation on various DNN benchmarks commonly used in existing DNNs, each of which is a \lax program.
These benchmarks provide the most fine-grained way to compare the performance of \sys and existing systems.
\Cref{tab:benchmarks} shows the six benchmarks in our evaluation.
GQA, RMSNorm, and GatedMLP are the main building blocks of large language models (LLMs).
%
QKNorm introduces query-key normalization before attention to enhance model convergence~\cite{chameleonteam2024chameleon}.
LoRA enables low-rank adaptation for fine-tuning a DNN on different tasks.
We use a context length of 8K for GQA and 4K for QKNorm, corresponding to the maximum supported by \llama-3-70B~\cite{grattafiori2024llama3herdmodels} and Chameleon-7B~\cite{chameleonteam2024chameleon}, respectively.
In addition, we also evaluate how \sys-generated kernels improve the end-to-end performance of full DNNs, including Chameleon~\cite{chameleonteam2024chameleon}, nGPT~\cite{loshchilov2024ngpt}, \llama-3~\cite{grattafiori2024llama3herdmodels}, and LoRA~\cite{hu2021lora}.

%
%

%
The experiments were conducted on NVIDIA A100 and H100 GPUs, each with 40GB of memory.
All our benchmarks fit on a single GPU except GQA (used for \llama-2-70B), which is generally parallelized across four GPUs using tensor model parallelism~\cite{Megatron}.
Therefore, we evaluate GQA under this parallelism strategy, where the eight key-value heads are equally partitioned across four GPUs.
Since the performance of \sys and all baselines depends only on the shapes of the input tensors, we repeat each experiment 1,000 times using random inputs and report the average run time. 

One of our benchmarks, LoRA, requires  concatenation to express a common optimization: fusing two matrix multiplications via concatenation. To support this optimization in \sys, we introduce a new linear operator that takes four inputs and computes $f(W, X, Y, Z) = (W \| X) \times (Y \| Z)$, where $\|$ is tensor concatenation.
This operator is equivalent to computing $W \times Y + X \times Z$.
We define the abstract expression associated with this operator as: 
$\expr(f(W, X, Y, Z)) = \eadd(\ered(k_1, \emul(\expr(W), \expr(Y))), \ered(k_2, \emul(\expr(X), \expr(Z))))$, where $k_1$ and $k_2$ are the last dimensions of $W$ and $X$.

Unless otherwise stated, \sys considers up to 5 operators in the kernel graph and up to 11 operators in each block graph.


\subsection{Benchmark Results}
\label{subsec:benchmark_results}
\Cref{fig:benchmarks} compares the performance of \sys with systems on six DNN benchmarks on NVIDIA A100 and H100 GPUs.
All systems use half-precision floating points to run these DNN benchmarks.
TASO~\cite{TASO} and PET~\cite{wang2021pet} are DNN superoptimizers that automatically generate algebraic transformations at the kernel level.
We report a combined TASO/PET baseline, as the latest TASO implementation includes PET’s partially equivalent transformations as special substitutions.
PyTorch~\cite{pytorch} uses the highly optimized cuDNN and cuBLAS libraries~\cite{cublas,cudnn} to perform DNN operators on GPUs.
For the PyTorch baseline, we enable {\tt torch.compile} and use FlashAttention kernels to maximize performance.
TensorRT and its LLM variant TensorRT-LLM include a set of manually designed and highly optimized kernels for common tensor operators such as attention~\cite{tensorrt}.
FlashAttention and its inference variant FlashDecoding are manually written kernels for efficient attention~\cite{dao2023flash, hong2024flashdecoding}.
Finally, Triton is a schedule-based optimizer to generate high-performance kernels and has been adopted in production systems, outperforming other schedule-based approaches~\cite{tillet2019triton}.
All baselines use CUDA Graphs to minimize kernel launch overhead.
%


Compared to the best existing approaches, \sys improves the performance of these benchmarks by up to $3.3\times$ by combing algebraic transformations, schedule transformations, and the generation of new custom kernels. \S\ref{sec:case2} shows the best discovered \graphs for RMSNorm. Next, we present a case study for the remaining benchmarks.

\begin{figure}
    \centering
    \subfloat[The kernel graph for QKNorm and attention in existing systems.]{
    \includegraphics[scale=0.37]{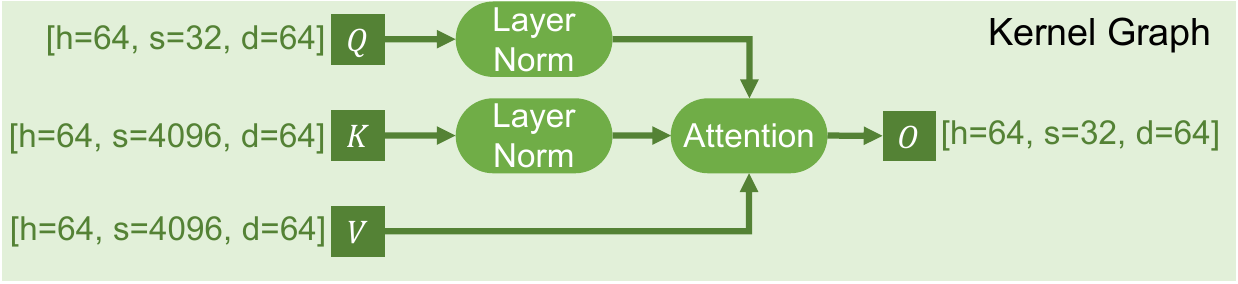}
    \label{fig:qknorm_baseline}
    }
    \\
    \subfloat[The best \graph discovered by \sys for QKNorm and attention.]{
    \includegraphics[scale=0.37]{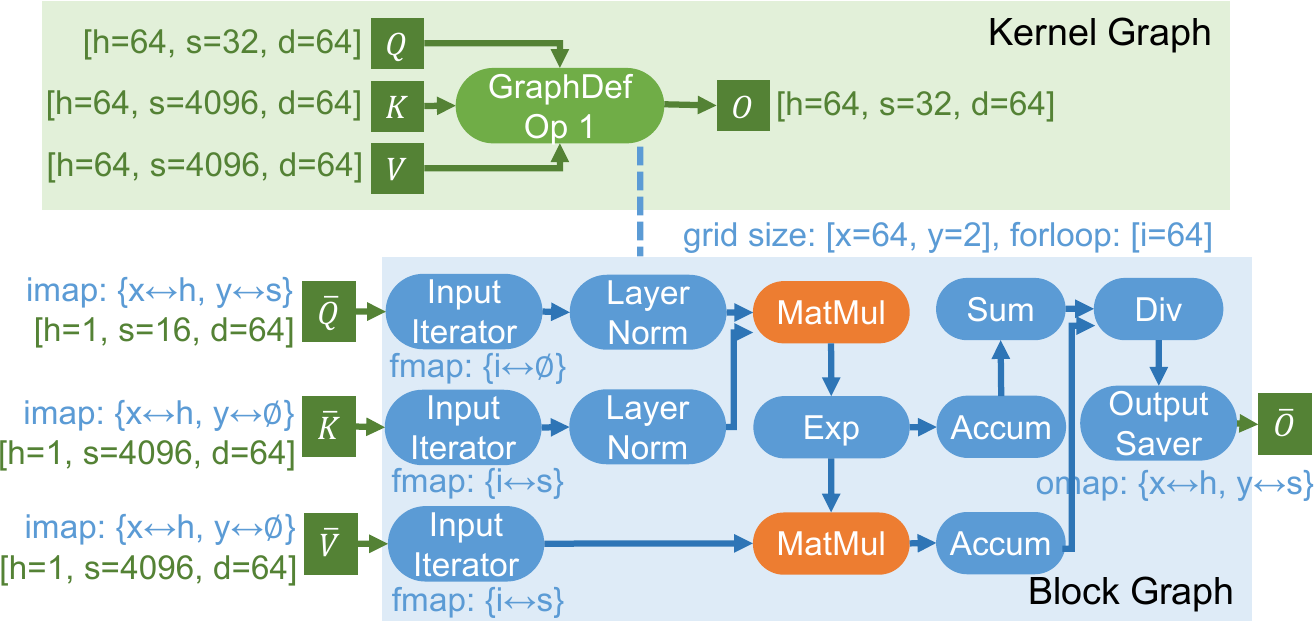}
    \label{fig:qknorm_ours}
    }
    \vspace{\captionvspace}
    \caption{Comparing the \graphs used by existing optimizers and \sys for QKNorm and attention.}
    \vspace{\captionvspace}
    \label{fig:qknorm_example}
\end{figure}

\paragraph{GQA.} Group-query attention is the backbone of LLMs and has been heavily optimized by existing frameworks. For example, FlashAttention and FlashDecoding are expert-designed attention kernels and have been adopted in existing LLM inference systems~\cite{dao2023flash}.
\sys discovers these expert-designed kernels as well as other \graphs that outperform them by up to 2.2$\times$.
The speedup is achieved by two additional optimizations on top of existing hand-written kernels. First, current approaches rely on fixed heuristics to determine the grid dimensions for GQA, which are suboptimal in certain scenarios. For example, TensorRT-LLM launches the GQA kernel with grid dimensions of (8, 2, 1) and (8, 2, 8) when the batch sizes are 1 and 8, respectively. However, both configurations cannot fully utilize all SMs on A100 (108 SMs) and H100 (132 SMs) GPUs. 
In contrast, \sys automatically searches for the best grid dimensions for each \graph, resulting in full SM utilization.
Further ablation study shows that the performance of the best \graph discovered by \sys degrades by 18\% when using the same grid dimensions as TensorRT-LLM.

Second, existing approaches use fixed tensor dimensions to parallelize GQA across thread blocks. For example, FlashAttention~\cite{dao2023flash} parallelizes attention across the {\em sample}, {\em head}, and {\em query sequence} dimensions, while FlashDecoding and TensorRT-LLM leverage the {\em sample}, {\em head}, and {\em key-value sequence} dimensions. Both strategies are efficient for conventional multi-head attention with many heads but suboptimal for GQA with fewer attention heads.
In contrast, \sys automatically selects the most efficient parallelization strategy by choosing among the sample, KV heads, query sequence, and key-value sequence dimensions. Moreover, \sys generates different \graphs tailored to different attention scenarios, reducing device memory access by up to 7$\times$ compared to the heuristics used in existing systems.
%
%

Implementing \sys's \graphs in existing systems is possible but requires extensive engineering effort to support different kernels for different scenarios. In contrast, \sys automatically generates them and verify their correctness.

\paragraph{QKNorm.} To reduce model divergence, several recent DNNs introduce query-key normalization (QKNorm) into the Transformer architecture~\cite{chameleonteam2024chameleon}. QKNorm applies layer normalization to the query and key vectors before attention, as shown in \Cref{fig:qknorm_baseline}. These additional normalization layers are not yet supported by existing attention implementations (e.g., FlashAttention and TensorRT-LLM) and require launching separate kernels for normalization and attention.

\Sys automatically discovers a \graph that integrates QKNorm and attention computation into a custom kernel, as shown in \Cref{fig:qknorm_ours}. The \graph reorganizes the attention computation to enable fusion with the two layer normalizations, which avoids writing intermediate results to GPU device memory and reduces the kernel execution time by up to 1.4$\times$.

\begin{figure}
    \centering
    \subfloat[The kernel graph for LoRA in existing systems.]{
    \includegraphics[scale=0.38]{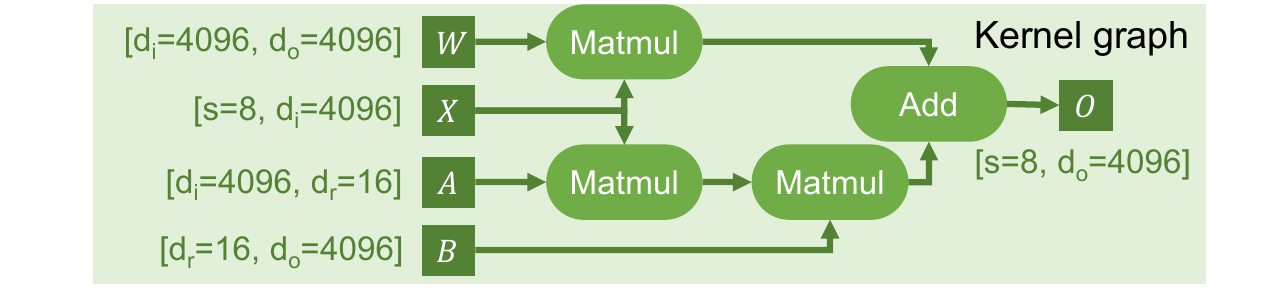}
    \label{fig:lora_baseline}
    }
    \\
    \subfloat[The best \graph discovered by \sys for LoRA.]{
    \includegraphics[scale=0.38]{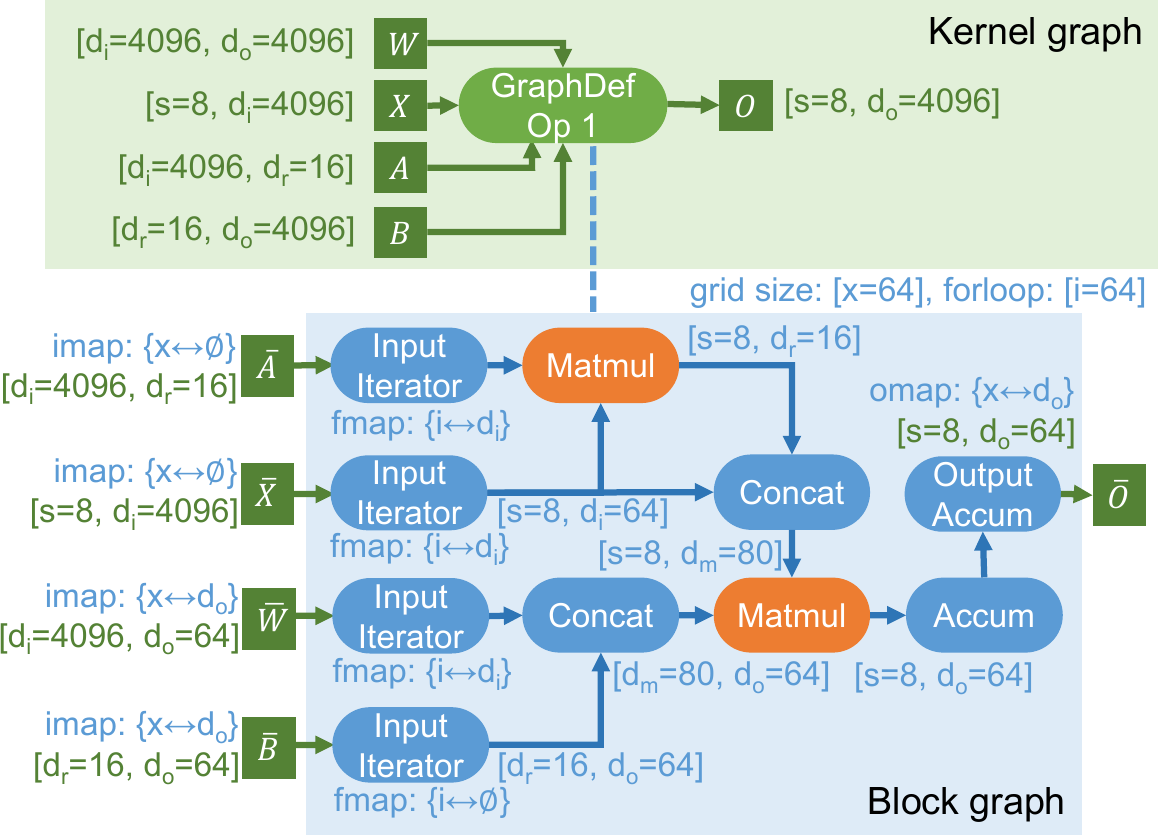}
    \label{fig:lora_ours}
    }
    \vspace{\captionvspace}
    \caption{Comparing the tensor programs used by existing optimizers and by \sys for LoRA: $O = W \times X + B \times A \times X$. Note that both matrices $A$ and $B$ are low-rank.}
    \vspace{\captionvspace}
    \label{fig:lora_example}
\end{figure}

\paragraph{LoRA.}
Low-rank adaptation (LoRA) introduces a pair of low-rank adapters to the linear operators of a pre-trained DNN to improve its performance for downstream tasks. 
Existing tensor program optimizers launch separate kernels for the original linear operator and the two additional linear operators introduced by LoRA (\Cref{fig:lora_baseline}), which introduces high kernel launch overheads since these LoRA operators involve minimal computation.
\Cref{fig:lora_ours} shows the best \graph discovered by \sys for LoRA, which fuses the three {\tt Matmul}s and the subsequent {\tt Add} into a single kernel. \sys reorganizes the computation into two block-level {\tt Matmul}s by leveraging the following algebraic transformation: $W \times X + B \times A \times X = (W \| B) \times \big(X \| (A\times X)\big)$. The {\tt Concat}s in \Cref{fig:lora_ours} do not involve any computation and are performed by updating tensor offsets in GPU shared memory.
This \graph reduces the execution cost of LoRA by
1.1-2.4$\times$.

\begin{figure}
    \centering
    \subfloat[The kernel graph for GatedMLP.]{
    \includegraphics[scale=0.4]{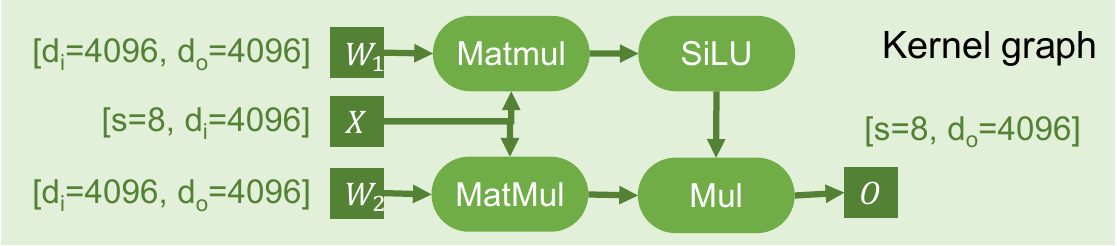}
    \label{fig:mlp_baseline}
    }
    \\
    \subfloat[The best \graph discovered by \sys for GatedMLP.]{
    \includegraphics[scale=0.4]{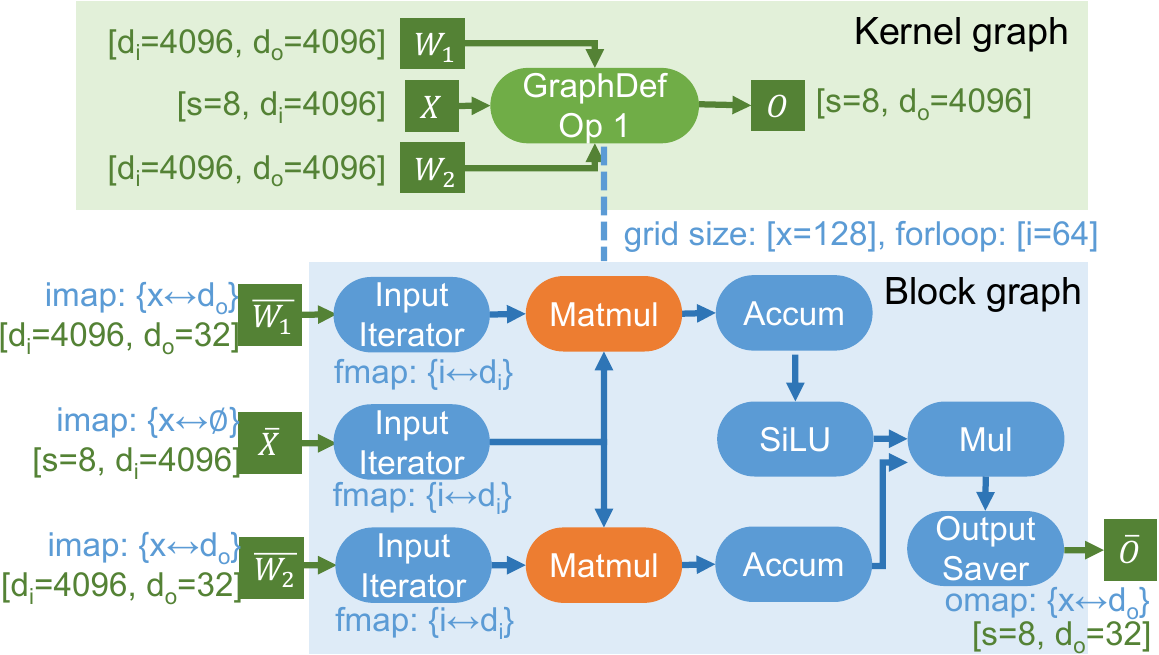}
    \label{fig:mlp_ours}
    }
    \vspace{\captionvspace}
    \caption{Comparing the \graphs used by existing optimizers and \sys for GatedMLP.}
    \vspace{\captionvspace}
    \label{fig:mlp_example}
\end{figure}

\paragraph{GatedMLP.}
Gated multi-layer perceptrons are commonly used in DNNs to capture non-linear representations.
We use the GatedMLP configuration introduced in Falcon-7B~\cite{falcon40b}, whose kernel graph is shown in \Cref{fig:mlp_baseline}.
Existing tensor program optimizers generally fuse the two {\tt Matmul}s in a single kernel to reduce GPU device memory access, since the input tensor $X$ only needs to be loaded once.
However, this approach still requires launching multiple kernels and storing intermediate results---specifically, the output of the two {\tt Matmul}s---in device memory, as the {\tt SiLU} activation and elementwise multiplication are not fused with the {\tt Matmul}s. 

In contrast, the best \graph discovered by \sys (\Cref{fig:mlp_ours}) performs the two {\tt Matmul}s in parallel within the same block graph and fuses the remaining computation (i.e., {\tt SiLU} and {\tt Mul}) as post-processing steps within the same block graph. This approach yields 1.5$\times$ speedups on A100 GPUs and 2.7-3.3$\times$ speedups on H100 GPUs.

%

\paragraph{nTrans.}
To accelerate model training, nGPT introduces normalized Transformer, which normalizes all intermediate results in Transformer~\cite{loshchilov2024ngpt}. Formally, the computation is defined as $y = \texttt{Norm}(x + \alpha (\texttt{Norm}(h - x)))$, where {\tt Norm} is a normalization layer, and $x$, $h$, and $\alpha$ are input tensors. Existing systems launch three separate kernels for nTrans, since it interleaves normalization and elementwise addition and multiplication. \Sys automatically discovers a \graph that fuses the computation into a single kernel and stores all intermediate results in GPU shared memory. \Sys outperforms other baselines but is slower than TensorRT. This performance gap is because \Sys loads data from global memory to shared memory and writes it back for each tensor in graph-defined kernels. This design improves memory efficiency and enables asynchronous pipelines. However, for kernels with light computation, the overhead of these memory transfers can dominate the kernel runtime. 
To mitigate this overhead, we plan to extend \sys to support bypassing shared memory during data loading, therefore avoiding unnecessary data movement.

\begin{figure}
    \centering
    \includegraphics[width=\columnwidth]{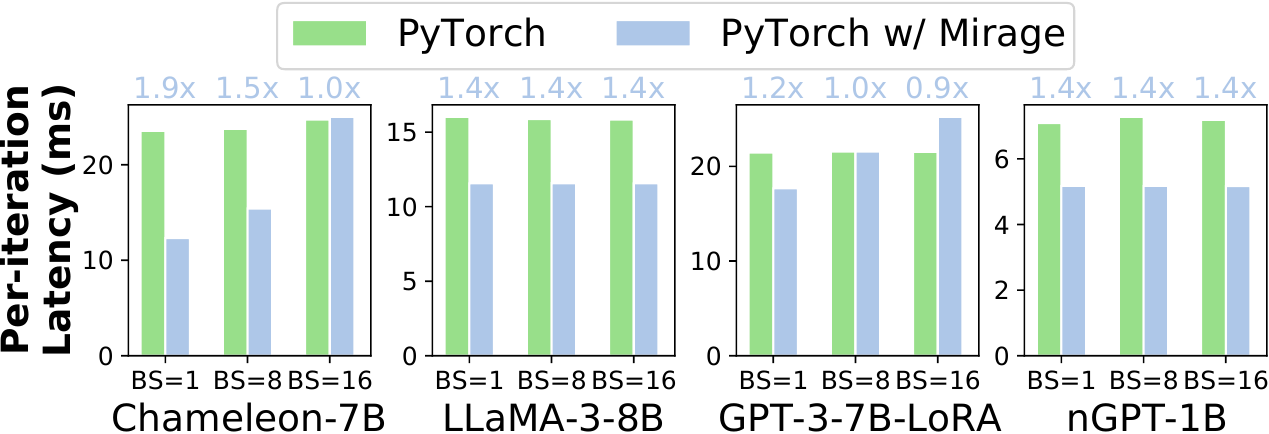}
    \caption{Comparing the end-to-end inference performance of PyTorch and PyTorch with \sys-generated kernels.}
    \label{fig:end_to_end_models}
\end{figure}

\subsection{End-to-end Results}
In addition to the microbenchmark performance, we also evaluate how \sys-generated kernels impact the end-to-end latency of commonly used DNNs. \sys supports just-in-time compilation and deployment, and its generated kernels can be directly integrated into PyTorch programs. We compare PyTorch with its native handwritten CUDA kernels and PyTorch with \sys-generated kernels on four DNN models. \Cref{fig:end_to_end_models} shows the results. \sys reduces the end-to-end latency of these models by 0.9-1.9$\times$ by automatically generating highly optimized kernels. The improvement is achieved with a few lines of code changes to the PyTorch programs.

\subsection{Search Time}

\begin{table}
    \centering
    \caption{Ablation study on \sys's techniques to accelerate \graph generation. We evaluate the impact of multi-threading and abstract expressions on search time for RMSNorm.}
    \vspace{\captionvspace}
    \footnotesize
    \label{tab:search_time}
    \begin{tabular}{c|r|r|r}
    \toprule
    {\bf Max \# Ops in} & {\bf \sys} & {\bf \sys w/o } & {\bf \sys w/o }\\
    {\bf a Block Graph} & {\bf } & {\bf Multithreading} & {\bf Abstract Expression} \\
    \midrule
    $5$ & $11$ sec &$58$ sec  & $768$ sec \\
    $6$ & $16$ sec & $93$ sec & $19934$ sec \\
    $7$ & $22$ sec & $150$ sec & $> 10$ h \\
    $8$ & $24$ sec & $152$ sec & $> 10$ h \\
    $9$ & $26$ sec & $166$ sec & $> 10$ h \\
    $10$ & $26$ sec & $166$ sec & $> 10$ h \\
    $11$ & $28$ sec & $183$ sec & $> 10$ h \\
    \bottomrule
    \end{tabular}
    \vspace{\captionvspace}
\end{table}

In our evaluation, \sys takes up to 4 hours to optimize a \lax program. This optimization is a one-time cost before deployment on the target hardware.
This subsection provides detailed results and an ablation study of \sys's search procedure, focusing on how its techniques enable the exploration of large \graphs while maintaining low search time.
In particular, we evaluate the impact of two techniques: pruning via abstract expressions (\S\ref{subsec:abstract_expression}) and multi-threading.
\Cref{tab:search_time} reports the search times for RMSNorm as we vary the maximum number of operators allowed in a block graph.

Multi-threading significantly reduces the search time, while pruning via abstract expressions is crucial for the scalability of \sys.
Specifically, the pruning techniques allow \sys to explore \graphs whose block graphs can each have at most 11 operators, while disabling abstract expression pruning restricts \sys to handle block graphs with up to 6 operators within a 10-hour search window.
Note that discovering the optimized \graph for RMSNorm shown in \Cref{fig:mugraph} requires exploring block graphs with 11 operators.

\subsection{Ablation Study on Optimizations}

\begin{figure}
    \centering
    \includegraphics[width=\columnwidth]{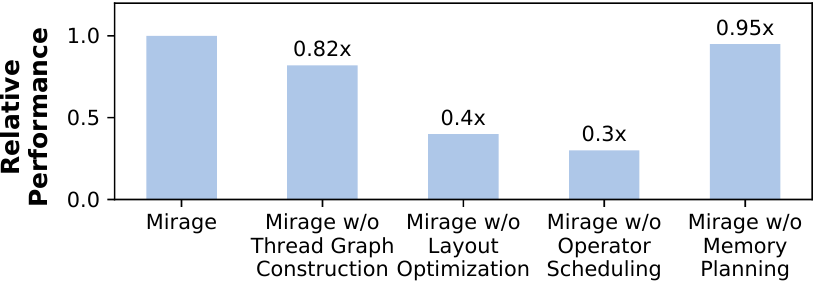}
    \vspace{\captionvspace}
    \caption{Ablation study on optimizations used in Mirage. We evaluate the performance degradation when disabling each optimization independently. The evaluation is performed on A100 for GQA with batch size $1$.}
    \vspace{\captionvspace}
    \label{fig:ablation-optimization}
\end{figure}

We conduct an ablation study to evaluate the impact of thread graph construction and optimizations introduced in \S\ref{sec:optimizer}, including layout optimization, operator scheduling, and memory planning. Specifically, we measure the performance degradation of the best \graph discovered by \sys when each optimization is disabled independently. The study is conducted on an A100 using the GQA benchmark with a batch size of $1$. The results, shown in \Cref{fig:ablation-optimization}, indicate that disabling any individual optimization leads to a performance degradation ranging from $5\%$ to $70\%$.

\section{Related Work}
\label{sec:related}

\paragraph{Manually-designed kernels.} Many existing frameworks, such as TensorFlow XLA~\cite{tensorflow_xla, Tensorflow}, PyTorch~\cite{pytorch}, and TensorRT~\cite{tensorrt}, rely on GPU experts to manually design kernels for ML operators.
Recently, significant engineering effort has been dedicated to hand-optimizing GPU kernels for commonly used DNNs, particularly foundation models~\cite{bommasani2022opportunities}.
For example, to accelerate attention computation~\cite{transformers}, several specialized kernels have been developed based on FlashAttention~\cite{dao2023flash, tri2023flashdecoding, hong2024flashdecoding, fastertransformer}. 
Due to the increasing complexity of modern GPUs---such as tensor cores in A100s~\cite{Markidis2018tensorcores} and thread block clusters in H100s~\cite{nvidia-h100}---manually designed kernels may miss subtle optimizations that are hard to discover manually.

\paragraph{Superoptimization-based approaches.} Superoptimization was originally introduced to find optimal instruction sequences~\cite{massalin, stoke, peephole}. 
Recent work has applied superoptimization techniques to tensor programs~\cite{TASO, wang2021pet, zheng2023einnet, tensat, unger2022unity, FlexFlow, hu2024korch, jeon2025graphpipe}.
However, all these attempts only consider algebraic transformations at the kernel level and cannot discover more sophisticated optimizations that require jointly considering algebraic and schedule transformations at all of the kernel, block, and thread levels.
Our evaluation shows that \sys largely outperforms existing DNN superoptimizers, demonstrating the importance of multi-level joint optimization.

\paragraph{Schedule-based approaches.} Recent work has introduced ML compilers that automatically optimize the execution schedule of kernel GPUs. Systems such as TVM~\cite{tvm, tvm_auto_tuner}, Ansor~\cite{ansor}, and Triton~\cite{tillet2019triton}, along with others~\cite{zheng2020flextensor, hagedorn2023graphene, feng2022tensorir}, build on the idea of algorithm-schedule separation introduced in Halide. They search for optimized schedules to execute a user-specified algorithm on GPUs.
However, schedule-based approaches require users to explicitly specify the algorithm for each kernel, and their performance is limited to the quality of these provided algorithms.

\paragraph{Multi-level graph representations.} Welder~\cite{shi2023welder} and ASPEN~\cite{park2023aspen} introduce multi-level tile graphs that share similarities with \sys's \graphs, as both representations follow the GPU hierarchy. However, prior work focuses on scheduling transformations, while \sys extends beyond scheduling by also considering algebraic transformations and the discovery of new custom kernels. Most optimizations presented in this paper fall outside the scope of these prior approaches.

\section{Conclusion}
\label{sec:conclusion}
\iffinal
This paper proposes \sys, the first multi-level superoptimizer for tensor programs. \sys introduces a hierarchy graph representation to specify a tensor program at the kernel, thread block, and thread levels of the GPU execution hierarchy, and uses a novel pruning technique based on abstraction to significantly reduce the search space \sys needs to consider while providing a certain optimality guarantee. \sys outperforms existing tensor program optimizers by up to 3.3$\times$, even for widely used and heavily optimized DNNs.
\else
This paper proposes \sys, the first multi-level superoptimizer for tensor programs. \sys introduces a hierarchy graph representation to specify a tensor program at the kernel, thread block, and thread levels, and uses a novel abstraction-based pruning technique to reduce the search space. \sys outperforms existing tensor program optimizers by 1.1-2.9$\times$ even for widely used and heavily optimized DNNs.
\fi 

\iffinal
\section*{Acknowledgment}
We would like to thank the anonymous reviewers and our shepherd, Stephanie Wang, for their valuable comments and suggestions.
We thank Tianqi Chen, Phillip Gibbons, Bohan Hou, Muyan Hu, Jinchen Jiang, Xiaoyu Jiang, Ruihang Lai, Yu Zhou, and other CMU Catalyst members for their feedback on this work. This research is partially supported by NSF awards CNS-2147909, CNS-2211882, and CNS-2239351, and research awards from Amazon, Cisco, Google, Meta, NVIDIA, Oracle, Qualcomm, and Samsung.
This research is also partially supported by a research grant from the Center for New Scientists at the Weizmann Institute of Science and by a grant from the Azrieli Foundation.
\fi

\balance
\bibliography{bibliography}
\bibliographystyle{plain}


\end{document}